%% file: main.tex
\documentclass[10pt,journal,compsoc]{IEEEtran}
\IEEEoverridecommandlockouts
\usepackage{amsfonts}
\usepackage{amsthm,amsmath}
\usepackage{mathrsfs}
\usepackage{indentfirst}
\usepackage{cite}
\usepackage{amsmath,amssymb,amsfonts}
\usepackage{algorithmicx}
\usepackage{graphicx}
\usepackage{textcomp}
\usepackage{xcolor}
\usepackage{booktabs} 
\usepackage[linesnumbered,ruled]{algorithm2e} 
\usepackage{amssymb,amsfonts}
\usepackage{subfigure}
\usepackage{makecell}
\usepackage{multirow}
\usepackage{graphicx,epstopdf}
\usepackage{subfloat}
\usepackage{url}
\usepackage{algpseudocode}
\usepackage{mdwlist}
\usepackage{cases}
\usepackage{bm}
\usepackage{enumitem}
\usepackage{subfigmat}
\usepackage{tabularx}
\usepackage{wrapfig}
\usepackage{xcolor}

\newtheorem{definition}{Definition}

\begin{document}

\title{Learning to Help Emergency Vehicles Arrive Faster: A Cooperative Vehicle-Road Scheduling Approach\\
}
\author{Lige~Ding,~Dong~Zhao,~Zhaofeng~Wang,~Guang~Wang,~Chang~Tan,~Lei~Fan~and~Huadong~Ma,~\IEEEmembership{Fellow,~IEEE}
\IEEEcompsocitemizethanks{
\IEEEcompsocthanksitem L. Ding, D. Zhao, Z. Wang and H. Ma are with Beijing Key Laboratory of Intelligent Telecommunication Software and Multimedia, School of Computer Science, Beijing University of Posts and Telecommunications, Beijing 100876, China. \protect\\
E-mail:$\{$dinglige, dzhao, wangzhaofeng, mhd$\}$@bupt.edu.cn
\IEEEcompsocthanksitem
G. Wang is with Institute for Data, Systems, and Society, Massachusetts Institute of Technology, Cambridge 02139, USA. \protect\\
E-mail:guangw@mit.edu
\IEEEcompsocthanksitem
C. Tan and L. Fan are with iFLYTEK Research, IFLYTEK Co., Ltd, Hefei, Anhui 230088, China. \protect\\
E-mail:$\{$changtan2, leifan$\}$@iflytek.com}
}


\IEEEtitleabstractindextext{
\input{abstract}

}

\maketitle
\input{1introduction_v1}

\input{2motivation}
\input{3preliminary}

\input{4method}

\input{6evaluation}
\input{8relatedwork}
\input{9conclusion}
\bibliographystyle{IEEEtran}
\bibliography{references}
\end{document}

%% file: abstract.tex
\begin{abstract}
The ever-increasing heavy traffic congestion potentially impedes the accessibility of emergency vehicles (EVs), resulting in detrimental impacts on critical services and even safety of people’s lives. Hence, it is significant to propose an efficient scheduling approach to help EVs arrive faster. 
Existing vehicle-centric scheduling approaches aim to recommend the optimal paths for EVs based on the current traffic status while the road-centric scheduling approaches aim to improve the traffic condition and assign a higher priority for EVs to pass an intersection. 
With the intuition that real-time vehicle-road information interaction and strategy coordination can bring more benefits, we propose \emph{LEVID}, a \underline{LE}arning-based cooperative \underline{V}eh\underline{I}cle-roa\underline{D} scheduling approach including a real-time route planning module and a collaborative traffic signal control module, which interact with each other and make decisions iteratively. 
The real-time route planning module adapts the artificial potential field method to address the real-time changes of traffic signals and avoid falling into a local optimum.
The collaborative traffic signal control module
leverages a graph attention reinforcement learning framework to extract the latent features of different intersections and abstract their interplay to learn cooperative policies.
Extensive experiments based on multiple real-world datasets show that our approach outperforms the state-of-the-art baselines.
\end{abstract}

%% file: 1introduction_v1.tex
\section{Introduction}
\label{sec:1introduction}
With the continual growth of population and vehicles in cities, we have been facing increasingly serious traffic congestion.
Heavy traffic congestion not only causes extra air pollution and energy/time waste, but also potentially impedes the accessibility of Emergency Vehicles (EVs), such as ambulances, fire engines and police cars, when facing unexpected accidents, resulting in detrimental impacts on critical services and even safety of people's lives.
In medical emergencies such as cardiac arrest, every one-minute delay causes mortality rate to increase by $1\%$ and imposes additional \$1542 medical cost in USA \cite{rapidsos}.
The building fires typically grow by $20\%$ per minute, causing an average \$4000 of additional damages \cite{rapidsos}.
Therefore, it is of great significance to design an efficient scheduling approach to help EVs arrive faster, especially in congested traffic conditions.

Different from Ordinary Vehicles (OVs), EVs may be exempted from some conventional road rules, such as driving through an intersection when the traffic light is red, or exceeding the speed limit. Nevertheless, EVs may still be obstructed by numerous OVs on roads with a heavy traffic.
To address this issue, one research line resorts to \textit{vehicle-centric} approaches, which aim at scheduling EVs with the best routes using route optimisation techniques such as the A* algorithm \cite{nordin2012finding}, Dijkstra’s algorithm \cite{chen2014path} and evolution strategy \cite{barrachina2014reducing}.
Some of studies on route planning for OVs, which fall into two categories, \textit{trajectory-based} approaches \cite{yuan2010t, dai2015personalized, guo2018learning, yang2020learning} and \textit{cost-centric} approaches \cite{hu2018risk,  yuan2019weight, li2019time, pedersen2020fast, yang2018pace}, 
could also be adapted to address the EV routing problem.
However, the vehicle-centric approaches just avoid congested roads in a passive way, while failing to proactively improve traffic conditions to shorten the travel time of EVs.
Another research line focuses on \textit{road-centric} approaches \cite{kang2014traffic,  younes2018efficient, rosayyan2020decentralized}, which aim at actively improving \emph{local} traffic conditions to help EVs pass intersections quickly by granting traffic signal priority.
However, these approaches rarely consider the dynamic \emph{overall} traffic condition and the impact of a scheduling strategy on OVs.
If we blindly keep the traffic light green for EVs arriving at intersections, the traffic congestion may not be effectively alleviated, and even traffic flows in other directions may be obstructed, thus in turn causing a greater negative impact on the overall traffic condition and also EVs.

Recent years have witnessed the great advance in Cooperative Vehicle-Infrastructure Systems (CVIS), wherein the sensing infrastructure (e.g., cameras, GPS) monitors traffic conditions and vehicles' locations in real time, and the communication infrastructure enables vehicles and road infrastructure to exchange real-time information \cite{Danlaw}.
It provides a new opportunity to design a \textit{cooperative vehicle-road scheduling} approach.
Along this research line, we aim to dynamically optimize the route and concomitantly coordinate the traffic signals along the dynamically updated path for better handling the dynamic traffic flow.
However, it is a very challenging task as the route planning and traffic signal control have complex interactions as follows:

\begin{itemize}[leftmargin=1em,itemindent=0pt,listparindent=0pt]
\item
\textbf{Impact of frequently changing traffic signals on real-time route planning.}
The real-time route planning needs to consider not only dynamic traffic flows over the road network but also frequently changing traffic signals.
Some cost-centric approaches \cite{wei2012constructing, chen2016learning, liu2020polestar} are able to address dynamic traffic flows by a stochastic graph based on historical data.
However, the traffic signal changes are so frequent (minute-level) that it is hard to accurately predict the travel time costs of different routes with only historical data.

\item
\textbf{Collaborative traffic signal control based on dynamic routing.}
Firstly, the dynamic route planning causes different intersections to become upstream and downstream intersections of EVs, which has different influences on EVs according to the Kinenmatic-wave theory \cite{hayes1970kinematic}. It is significant but difficult to extract the latent features and dynamic influences of these intersections.
Secondly, multiple traffic lights should learn to cooperate   with each other to balance the traffic demands of both EVs and OVs.
The joint optimization may lead to the scale expansion of the problem and increase the computational complexity. Although extensive studies focus on traffic signal control for OVs, they cannot well handle our problem \cite{sommer2010bidirectionally,xu2019exploring,wang2020stmarl}. 
\end{itemize}

To this end, we propose \emph{LEVID}, a \underline{LE}arning-based cooperative \underline{V}eh\underline{I}cle-roa\underline{D} scheduling approach, consisting of a real-time route planning module and a collaborative traffic signal control module, which influence each other and make decisions iteratively.
The real-time route planning module adapts the artificial potential field method to address the real-time changes of traffic signals and avoid falling into a local optimum by considering the long-term cumulative benefit of a route.
The traffic signal control module leverages a graph attention reinforcement learning framework, which models the traffic environment as a dynamic directed graph to present the influences of dynamic routes and increase the receptive field of each agent (traffic signal controller). By employing the multi-head attention as relation kernel, this framework is able to extract the latent features of different intersections and abstract their interplay to learn cooperative policies.
Meanwhile, the asynchronous parameter-sharing method is adopted to reduce the computational complexity. Specifically, our contributions are three-fold as follows:

\begin{itemize}[leftmargin=1em,itemindent=0pt,listparindent=0pt]
\item
We investigate the \textit{cooperative vehicle-road scheduling} paradigm for helping EVs arrive faster. 
To the best of our knowledge, this is the first work to simultaneously optimize the route planning and traffic signal control in real time (Sect. \ref{sec:2motivation}, Sect. \ref{sec:3preliminary}).
\item
We propose the \emph{LEVID} approach, which considers the long-term cumulative benefit of a dynamically planned route and leverages graph attention reinforcement learning for better cooperation between neighboring intersections (Sect. \ref{sec:4method}).
\item
We evaluate our \emph{LEVID} using both synthetic and real-world datasets from multiple cities. Experimental results demonstrate that our approach greatly reduces the average travel time for both EVs and OVs than the state-of-the-art baselines (Sect. \ref{sec:6evaluation}). 
\end{itemize}

%% file: 2motivation.tex
\section{Motivation}
\label{sec:2motivation}

\begin{figure} 
\centering
\includegraphics[width=9.5cm]{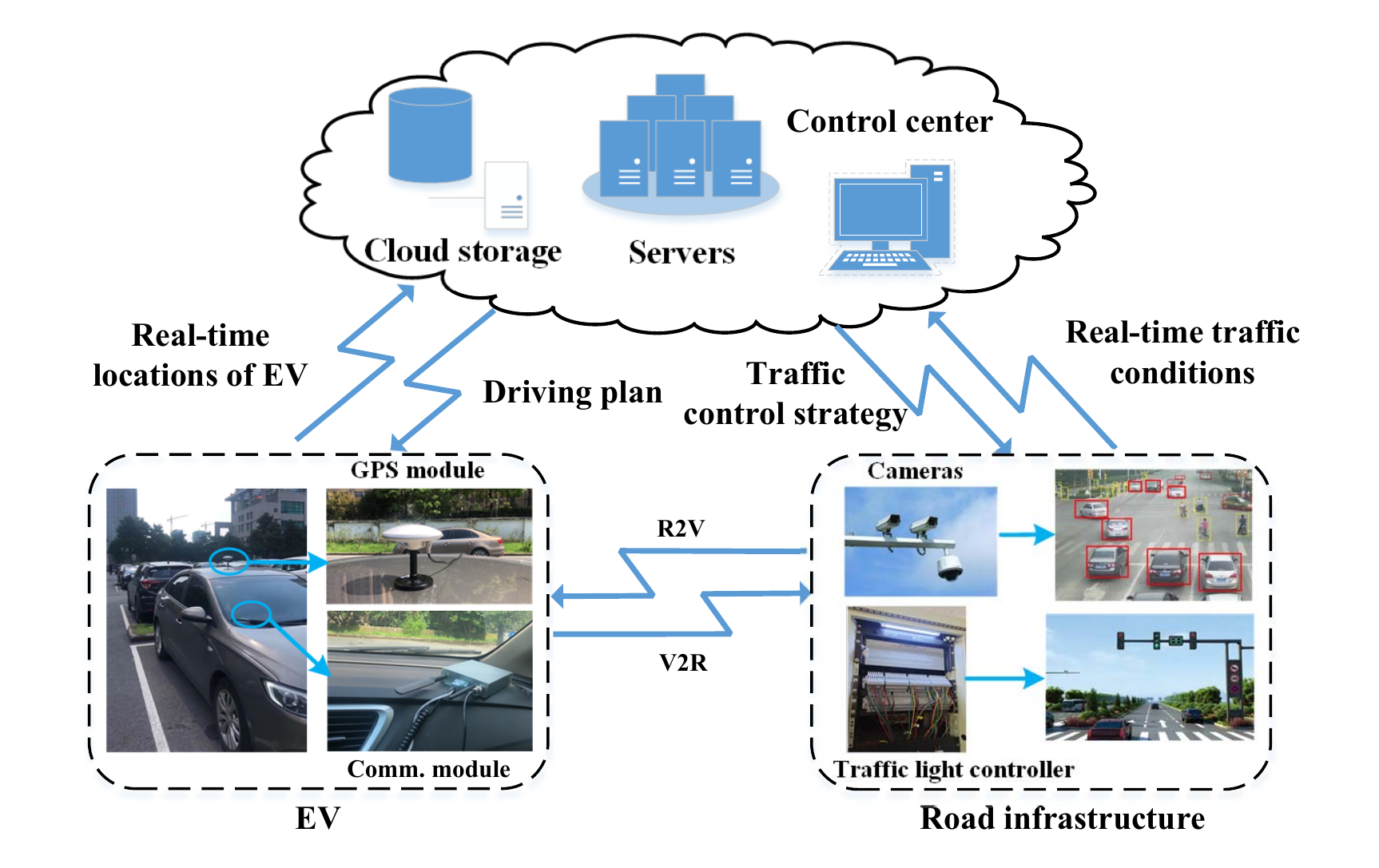}\vspace{-10pt}
\caption{The architecture of a CVIS.}
\vspace{-10pt}
\label{fig_CVIS}
\end{figure}
In this section, we first introduce the supporting devices and technologies of a CVIS, which provides opportunities for designing effective scheduling approaches. Second, we point out the defects of the existing approaches from two separated perspectives, i.e., traffic signal control and route planning, which motivates us to design a cooperative vehicle-road scheduling approach \textit{LEVID} for sufficiently leveraging the ability of a CVIS.


\textbf{CVIS.}
Fig. \ref{fig_CVIS} shows the architecture of a CVIS, which consists of EVs, road infrastructure and a control center.
On the EV side, a GPS module is used to collect real-time locations of an EV; a communication module is used to interact with the road infrastructure via Vehicle-to-Roadside (V2R) and Roadside-to-Vehicle (R2V) communications according to the Dedicated Short Range Communication (DSRC) standard, and also interact with the control center via the cellular communications (e.g., 4G/5G).
On the road infrastructure side, traffic cameras and traffic signal controllers have been widely deployed on major roads of many cities.
For example, there are over 3,000 major intersections in the urban area of Hefei city, China, of which 1,338 intersections have traffic signal controllers that can be adjusted by the control center, and there are 14,967 traffic cameras deployed at intersections and other locations such as entrances/exits of expressways and key locations along arterial roads, as partly shown in Fig. 2.
The trajectories of all the vehicles are recorded when they pass through cameras, and can be extracted by the advanced vehicle identification technologies \cite{tong2021large}. The traffic volume can also be obtained by counting the number of vehicles passing through intersections.
Finally, the control center can obtain real-time locations of EVs and traffic conditions from the road infrastructure; in turn, it can provide a driving plan to the EV and determine a traffic control strategy to the traffic signal controller.

\begin{figure}[t]
    \centering
        \begin{minipage}[t]{0.29\textwidth}
        \centering
        \includegraphics[height=3.1 cm]{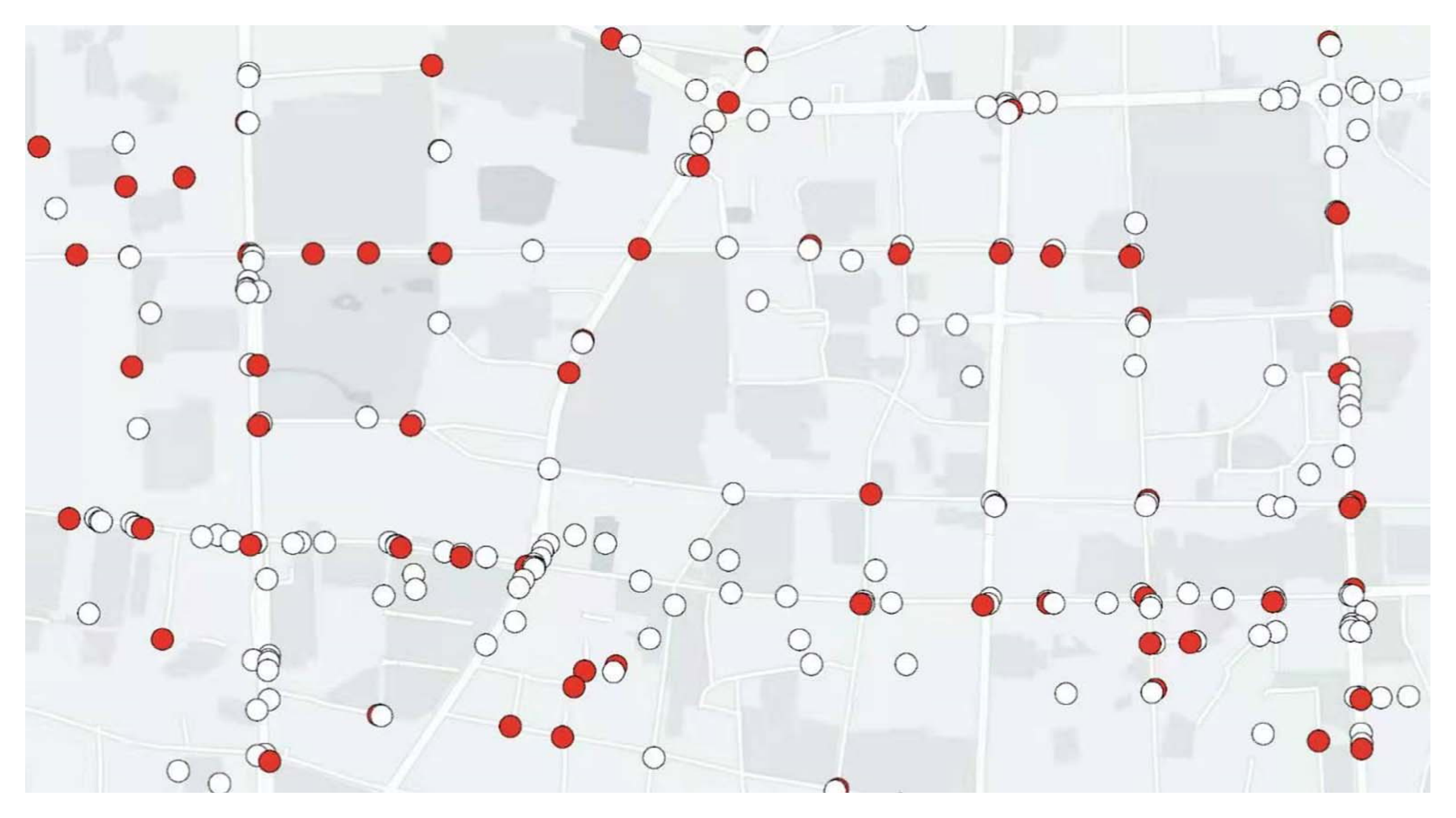}\vspace{-15pt}
        \label{fig_light_map}
        \caption{Distribution of cameras (denoted by white dots) and traffic signal controllers (denoted by red dots).}
    \end{minipage}
    \hfill
    \begin{minipage}[t]{0.17\textwidth}
        \centering
        \includegraphics[height=3.1 cm]{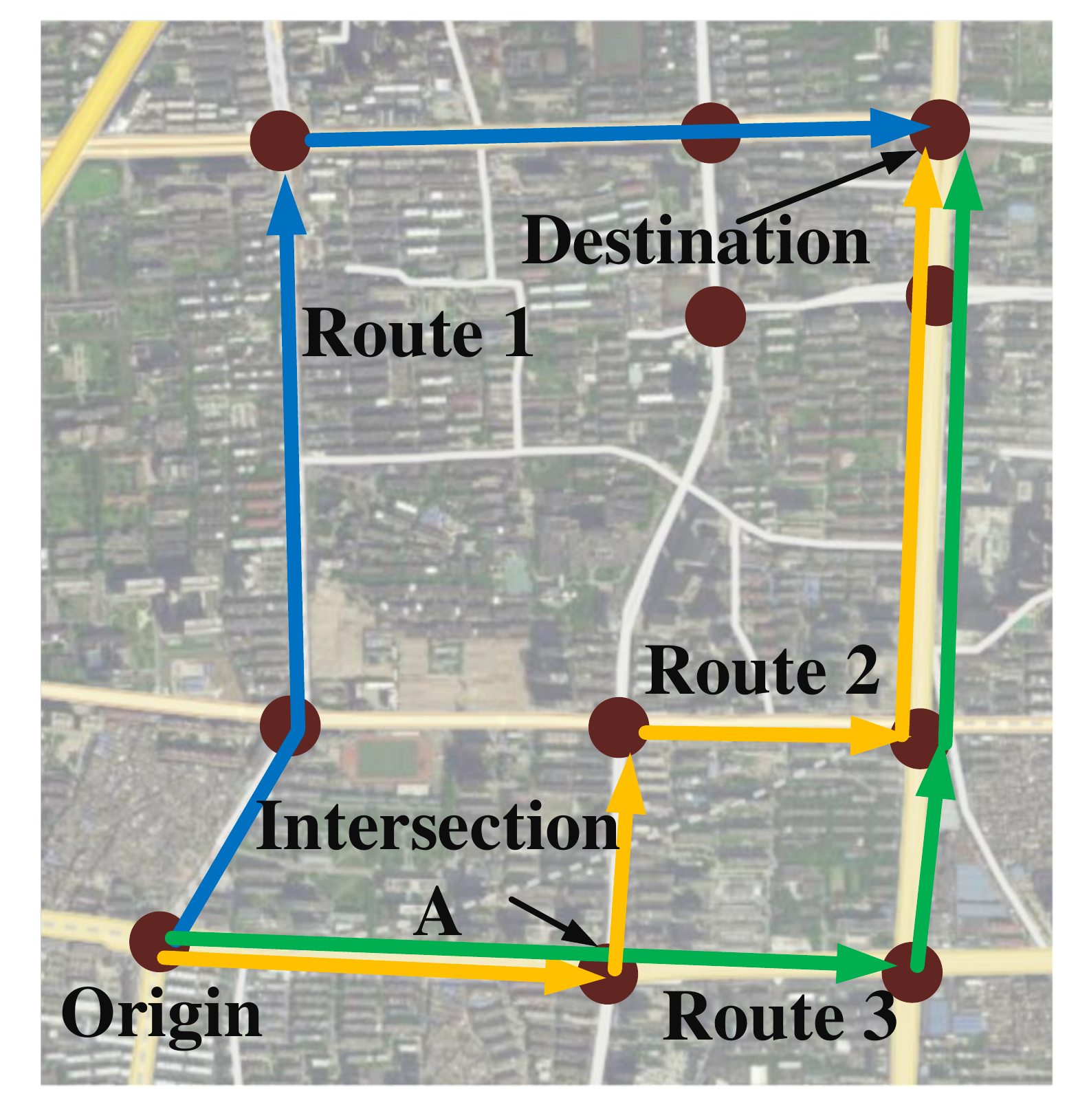}\vspace{-15pt}
        \label{fig_hefei_realmap}
        \caption{Three candidate routes with the given origin and destination in Hefei city.}
    \end{minipage}\vspace{-15pt}
\end{figure}

\textbf{Traffic signal control.} 
\emph{GreenWave} \cite{kang2014traffic} is the most commonly used traffic signal control approach, which allows all the traffic lights in the route to turn green so that EVs can pass intersections continuously along the emergency corridor. The ``green wave'' is achieved by signal coordination setting. 
However, if there is a traffic jam on one road segment, the signal offset time between intersections will be changed such that EVs cannot pass intersections continuously. 
In other words, \emph{GreenWave} cannot handle a dynamic and heavy traffic flow well.
To address this issue, we are working with the traffic police department in Hefei city to improve the \emph{GreenWave}.
More specifically, since the real-time locations of an EV are available, we can adjust the traffic signal to turn green automatically whenever the distance between EV and intersection is less than a certain threshold.
Nevertheless, it is still a non-trivial task to determine a proper threshold.
If the threshold is too large, it may cause the vehicles in the opposite direction to be blocked for a long time.
Conversely, if the threshold is too small, it may fail to clear the way for the EV at an congested intersection. 
To analyze this phenomenon, we collect the trajectory data of 5,448 vehicles from traffic cameras during 9-11 a.m. on one working day in a region of Hefei city (Fig. 3). We simulate the movement of an EV through an intersection.
When the threshold is set as $500m$ for an intersection with a low traffic pressure, the traffic flow in the opposite direction has to wait for extra $32$ seconds.
When the threshold is set as $200m$ for an intersection with a heavy traffic pressure, the EV is blocked by queuing vehicles at the intersection for about $47$ seconds. 
\textbf{This phenomenon motivates us to design a more effective approach from two perspectives: 1) utilize a learning-based traffic signal control strategy instead of a rule-based strategy, and 2) integrate it with a route planning strategy to further reduce the waiting time at intersections with a heavy traffic.}

\textbf{Route planning.}
To preliminarily demonstrate the importance of route planning, we generate an EV to move along different routes by data-driven simulations.
As shown in Fig. 3, given the same origin and destination, if the EV moves along \textit{Route 1} with the shortest distance, the travel time is $236.7s$; if it moves along \textit{Route 2} with the least congestion, the travel time would be $214.9s$; whereas, when we further consider the real-time status of traffic signals into account, \textit{Route 3} is the best choice with the travel time of $198.5s$, as the phase in the east-west direction is allowed and the left-turn phase is forbidden at \textit{Intersection A}.
\textbf{It implies the significance of considering both traffic conditions and changes of traffic signals for route planning.}

%% file: 3preliminary.tex
\section{Problem Formulation}
\label{sec:3preliminary}
\begin{definition}[Road Network] The road network is defined as a directed graph $G=(V, E)$, where $V=\{v_1, v_2, \cdots, v_I\}$ is the set of nodes (i.e., intersections) and $E$ is the set of edges (i.e., road segments). An edge $e_{i,j}\in E$ represents a directed road segment from intersection $v_i$ to intersection $v_j$.
\end{definition}

\begin{definition}[Route]
A route ${R}$ connects the origin location $v_o$ and the destination location $v_d$ with an ordered sequence of intersections, i.e., ${R}:v_o \to v_1 \to \cdots \to v_i \to \cdots \to v_d$, where each pair of consecutive locations corresponds to a road segment $e_{i,i+1}$. 
\end{definition}

\begin{definition}[Incoming/Outgoing Lanes and Traffic Movement]
For a specific intersection, we define that (i) a lane where vehicles enter the intersection is called as an incoming lane; (ii) a lane where vehicles leave the intersection is called as an outgoing lane; (iii) the traffic traveling across the intersection from an incoming lane $l$ to an outgoing lane $l'$ is called as a traffic movement, denoted by $(l,l')$. Each road segment contains one or multiple lanes. The sets of incoming lanes and outgoing lanes are denoted by $L_{in}$ and $L_{out}$.
\end{definition}

\begin{definition}[Movement Signal and Phase]
A movement signal $g(l,l')$ is defined based on the corresponding traffic movement $(l,l')$. 
Specifically, $g(l,l') = 1$ indicates that the green light is on for movement $(l,l')$, and $g(l,l') = 0$ indicates that the red light is on for movement $(l,l')$. A phase is defined as a combination of the legal green movement signals, denoted by $p=\{ (l,l')|g(l,l')=1\}$, where $l \in L_{in}$ and $l' \in L_{out}$.
\end{definition}

\begin{figure} 
\centering
\includegraphics[height=6 cm,width=8.5cm]{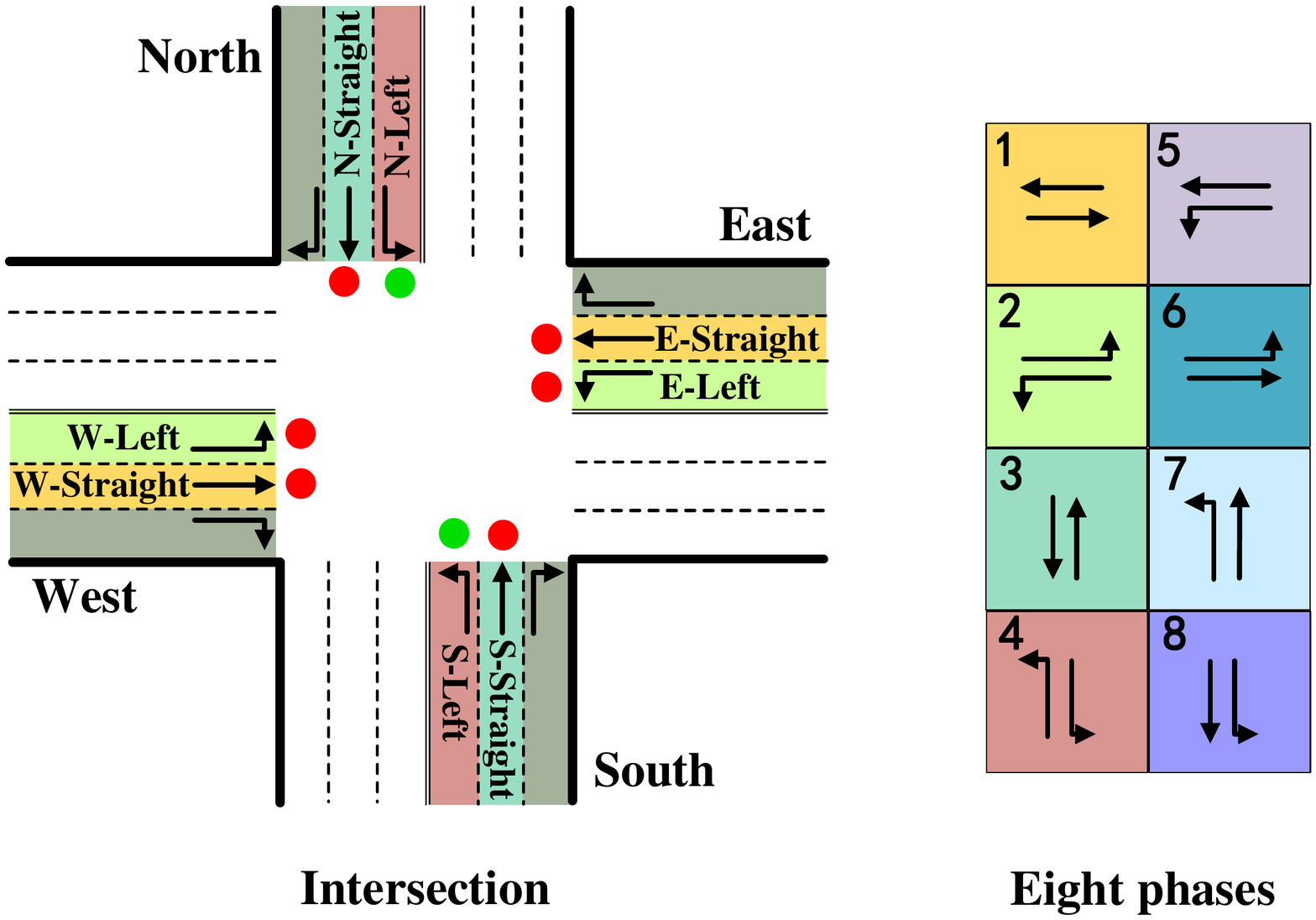}  \vspace{-15pt}
\caption{Illustration of an intersection with eight mutually exclusive phases. In this case, phase $\#4$ is activated to allow the S-Left and N-left traffic movements.}
\vspace{-12pt}
\label{phase}
\vspace{-10pt}
\end{figure}

\begin{figure*} [htbp] 
\centering
\includegraphics[width=17.5cm]{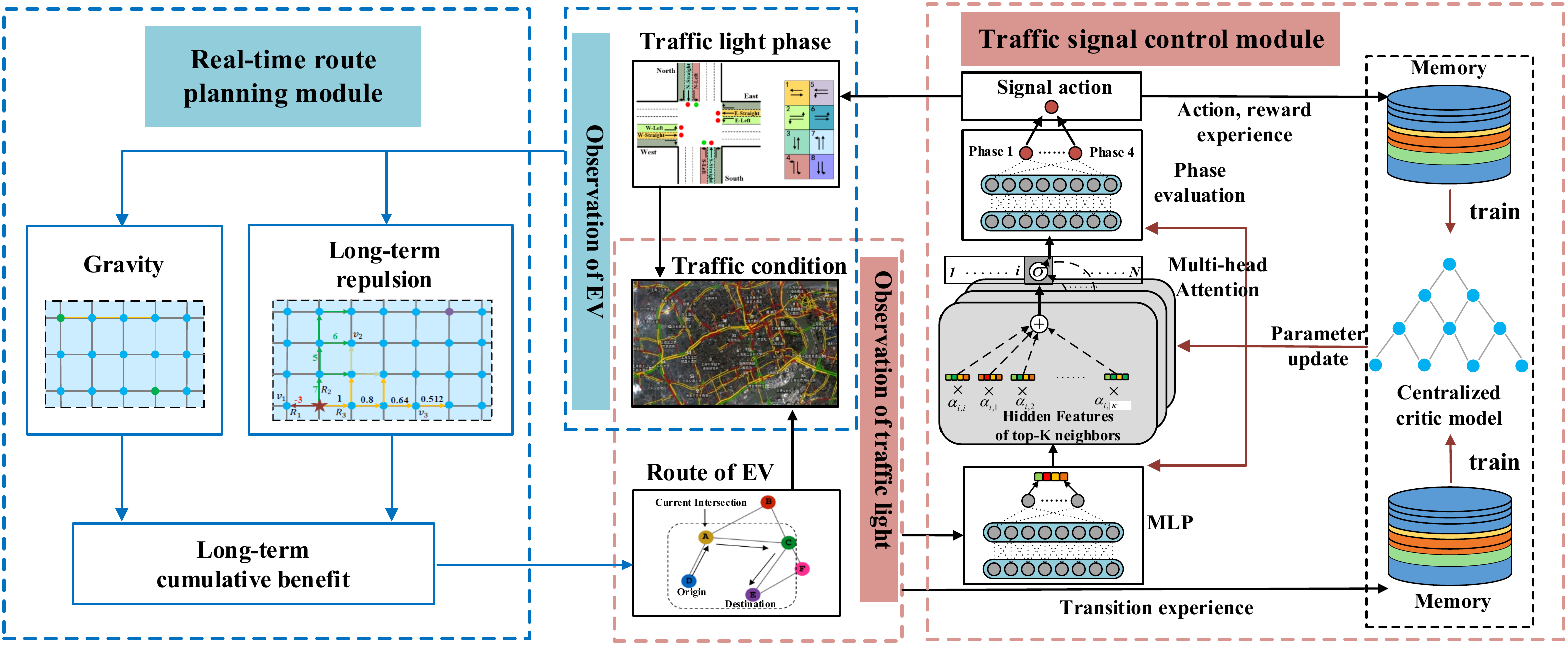}
\caption{Framework of \textit{LEVID}}
\label{Overview}
\vspace{-15pt}
\end{figure*}

Fig. \ref{phase} illustrates a typical intersection with twelve incoming lanes and twelve outgoing lanes. 
Correspondingly, there are eight movement signals (red and green dots around the intersection) for controlling traffic movements: \emph{E-Straight} (Go Straight from East), \emph{W-Straight}, \emph{S-Straight}, \emph{N-Straight}, \emph{E-Left} (Turn Left from East), \emph{W-Left}, \emph{S-Left}, and \emph{N-Left}.
Four right-turn signals are omitted as the traffics on the right-turn lanes are always allowed in the real world.
Furthermore, there are eight mutually exclusive phases, each of which is a combination of two traffic movements. 
In this example, the phase $\#4$ is activated, indicating that the traffics on the left-turn lanes from south and north are allowed to turn left.

\begin{definition}[Travel time]
Given a route ${R}$ of an EV, its travel time $T$ consists of the driving time $T_e$ on each road segment $e\in R$ and the waiting time $T_v$ at each intersection $v\in R$ to wait for the existing queued OVs to pass through the intersection. Note that, although an EV may be exempted from some conventional road rules, such as driving through an intersection when the traffic light is red, or exceeding the speed limit, it may still be obstructed by OVs on roads with a heavy traffic.
We denote the travel time of one EV by:
\begin{equation}
T({R}) = \sum_{e\in{R}}{T_e} + \sum_{v \in {R}}{T_v}.
\end{equation}
\end{definition}

\textbf{Problem Statement}. Given the origin location $v_o$ and the destination location $v_d$ of an EV and the dynamic traffic condition ${C}_t$ at each time step $t$, a real-time route planning strategy $\varphi_1$ is utilized to determine a driving route ${R}$.
Meanwhile, given the observation $o_i^t$, such as vehicle distribution and current traffic signal phase, of each intersection $v_i$ at each time step $t$, a collaborative traffic signal control strategy $\varphi_2$ is utilized to choose a control action (i.e., which phase to set).
The objective of this work is expressed as follows:
\begin{equation}
{\varphi^*} = \arg \min \limits_{\varphi_1,\varphi_2} T({R}|v_o,v_d,{C}_t).
\end{equation}

%% file: 4method.tex
\section{Design of \textit{LEVID}}
\label{sec:4method}

\subsection{Framework}
\label{subsec:framework}
As shown in Fig. \ref{Overview}, \textit{LEVID} contains a real-time route planning module and a collaborative traffic signal control module, which influence each other and make decisions iteratively.
\begin{itemize}[leftmargin=1em,itemindent=0pt,listparindent=0pt]
\item
\textbf{Real-time route planning module}
adapts the artificial potential field method \cite{khatib1985real} by modeling the estimated travel time as the repulsion and the trend of an EV moving towards the destination as the \emph{gravity}. Furthermore, it introduces the \emph{long-term repulsion} to handle the changing traffic lights and avoid falling into a local optimum. At every time interval $\vartriangle \!\! t$, the route with the maximum \emph{long-term cumulative benefit} is selected according to the current traffic signal phases and traffic condition near the EV. Meanwhile, the length of any candidate route is limited to reduce search depth and computational complexity. 

\item
\textbf{Traffic signal control module} models the traffic environment as a \emph{dynamic directed graph} and adjusts the relational distance between intersections according to the dynamically updated route and upstream/downstream relationships.
The receptive field of each agent contains its top-$K$ relevant neighboring intersections to differentiate valuable local information from global information. 
The observed features of the top-$K$ relevant intersections are transformed into hidden features with Multi-Layer Perceptron (MLP).
Then the \emph{multi-head attention} is employed as relation kernel to extract the latent features of different intersections and abstract their interplay to learn cooperative policies.
Finally, the long-term impacts of different traffic signal phases are evaluated by the \emph{centralized critic model} whose parameters are shared by all distributed actors (traffic signal controllers).
\end{itemize}

\subsection{Real-time route planning module}
\label{subsec:recommendation}



The detailed calculation process of gravity and long-term repulsion is introduced in the following part.

\textbf{Gravity.}
The gravity indicates the trend of an EV moving towards the destination.
The greater the gravity, the faster the EV can reach the destination.
Suppose an EV is arriving at the current intersection $v_c$, and will go to the final destination intersection $v_d$.
Then the gravity of $v_c$'s neighbor $v_i$ to the EV is calculated as:
\begin{equation}
F_g(v_c,v_i) = \big( dis(v_c,v_d) - dis(v_i,v_d) \big)/S_{c,i}
\end{equation}
where $dis(v_c,v_d)$ denotes the road network  distance between $v_c$ and $v_d$, $dis(v_i,v_d)$ denotes the road network  distance between $v_i$ and $v_d$, and $S_{c,i}$ is the real-time average traffic speed on the road $e_{c,i}$.

\textbf{Immediate Repulsion.}
The immediate repulsion represents the estimated travel time of a candidate route. It contains the driving time of an EV on the road segments and the waiting time at the intersections along the route. 
Suppose an EV is arriving at the current intersection $v_c$. 
Then the immediate repulsion of $v_c$'s neighbor $v_i$ to the EV contains time $t_w(v_c,v_i)$ to wait at the intersection $v_i$ and driving time $t_r(v_c,v_i)$ on the road segment $e_{c,i}$, calculated as follows:
\begin{equation}
\begin{aligned}
F_r(v_c,v_i) &=  t_r(v_c,v_i) + t_w(v_c,v_i)   \\   
&=  \frac{dis(v_c,v_i)\!-\!len(e_{c,i}) }{S_{c,i}} \!+ \!\frac{len(e_{c,i})}{ S'}     
\end{aligned}
\label{eq:fr}
\end{equation}  
where $len(e_{c,i})$ denotes the length of the queue about to drive from intersection $v_c$ to intersection $v_i$. $S_{c,i}$ is the real-time average traffic speed on the road $e_{c,i}$ and $S'$ denotes the maximum speed for vehicles passing through an intersection allowed by law. 

\textbf{Long-term Repulsion.}
Long-term repulsion helps approximate the long-term cumulative benefit of one route. Some routes with less immediate repulsion may guide vehicles to move to a congested road segment due to short-term shortcomings.
Therefore, we expand the search depth and calculate the long-term repulsion ${F'}_r(v_c,v_i)$ along different routes with a discounted factor $\lambda$ as follows:
\begin{equation}
\label{longtermF}
{F'}_r(v_c,v_i) = F_r(v_c,v_i) + {\lambda} \min_{v_j \in \mathcal{N}_i}{{F'}_r(v_i,v_j)}
\end{equation}
where $\mathcal{N}_i$ denotes the set of $v_i$'s neighbors, and this iterative calculation will stop when the search depth $num$ reaches the maximum search depth limit $Dep$.
The repulsion is approximated based on the current traffic condition and it may have changed when an EV travels to the relevant road segment far away from the current location.
The greater the distance between intersections, the larger the error of the estimated long-term impact. 
Therefore, a smaller discounted factor will be assigned to a farther intersection.
We limit the depth of search space and calculate the long-term discount repulsion according to Eq. (\ref{longtermF}).

\begin{figure} 
\centering
\includegraphics[width=8.5cm, height = 3.3 cm] {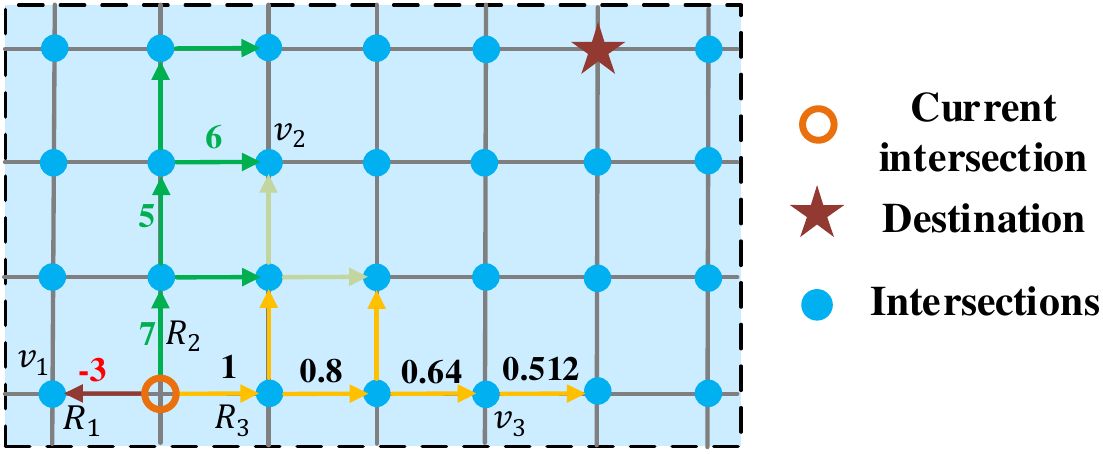}\vspace{-5pt}
\caption{The example of gravity and repulsion of different intersections. The red  number shows the gravity of intersection $v_1$ towards EV. The value is negative as it will lead EV away from destination. The green numbers denote the repulsion while the black numbers show the discount factors.}
\vspace{-8pt}
\label{Repulsion}
\end{figure}

Fig. \ref{Repulsion} illustrates the detailed process of route planning. 
The orange circle denotes the current location of one EV.
There are three candidate routes, i.e. $R_1$-red, $R_2$-green, $R_3$-yellow. Based on a specific traffic condition, the gravity of intersection $v_1$ towards EV is $-3$ as this intersection will lead the EV to move away from the destination.
The green numbers show the immediate repulsion of each road segment in route $R_2$.
Taking route $R_3$ as an example, we introduce the discount factor. As the distance increases, the discount factor decreases exponentially.
We limit the depth of search space to 4 and set the discount factor $\lambda$ as 0.8 in this example to show the detailed process.
The complete algorithm is shown in Algorithm \ref{alg:Route}.

\label{apx:alg}
\vspace{-5pt}
\begin{algorithm}
\SetAlgoLined
\caption{The real-time route planning module}
\label{alg:Route}
Input: Current location $v_c$ of EV, destination $v_d$, the state of each road segment.  \\
Initialize the discount factor $\lambda$, and the depth of search space $Dep$  \\
\For{$\vartriangle \!\! t$ $\in$ $T$}
{
\For{$v_i$ $\in$ $\mathcal{N}_c$}
{
/*Calculate $v_i$'s gravity towards EV*/
$F_g(v_c,v_i) \! = \! \big( dis_{m}(v_c,v_d) \! - \! dis_m(v_i,v_d) \big)/S_{c,i}$; \\
/*Calculate $v_i$'s immediate repulsion*/ \\
$F_r(v_c,v_i) =  t_r(v_c,v_i) \ +\ t_w(v_c,v_i) $; \\ 
/*Calculate $v_i$'s long-term repulsion*/\\
${F'}_r(v_c,v_i) \! = \!F_r(v_c,v_i) \! + \! {\lambda} \! \min_{v_j \in \mathcal{N}_i} \!{{F'}_r(v_i,v_j)}$; \\
/*Calculate long-term cumulative benefit*/ \\
$B(v_c,v_i) = F_g(v_c,v_i) - F'_r(v_c,v_i)$; \\
}
}
Output: The route with the maximum long-term cumulative benefit $B(v_c,v_i)$ \\
\end{algorithm}

\subsection{Traffic signal control module}
\label{subsec:signal_control}
The control of traffic signals can be formulated as decentralized partially observable markov decision process, where each agent chooses its phase action based on local observation $o^i$ at each time interval $\vartriangle \!\! t$.
\subsubsection{\textbf{Agent design}}
\label{subsubsec:agent_design}
The state, action and reward for an agent which controls the signal of one intersection are as follow:
 
\textbf{State (Observation).} State $\mathcal{S}$ denotes the traffic conditions of the whole urban environment while the observation of one agent in multi-agent RL equals to the state of the intersection. The observation $o^i$ of one agent at intersection $v_i$ includes the current phase $ph^i$, the number of OVs $x_o(l)$ on each entering lane $L_{in}^i$, the number of OVs $x_o(l')$ on each exiting lane  of this intersection and the corresponding number of EVs on each entering lane $L_{in}^i$ and exiting lane $L_{out}^i$, which are denoted as $x_s(l)$ and $x_s(l')$.

\textbf{Action.} At time $t$, each agent chooses different legal available phase set according to the structure of road network and traffic demand.
In our problem, we consider four phases (\emph{WE-Straight}, \emph{NS-Straight}, \emph{WE-Left} and \emph{NS-Left}) for an intersection.

\textbf{Reward.} 
The traffic light control method should consider both OVs and EVs. Therefore, we design the reward with an evaluation mechanism which considers these two types of traffic demands. We utilize the pressures to help OVs go through intersections more smoothly. As for EVs, they need to pass as soon as possible. Thus we leverage the queue length to measure the benefit of one action.
The pressure \cite{varaiya2013max, wei2019presslight} of a movement for OVs is defined as the difference of OV density between the entering lanes and the exiting lanes. The pressure $P_o(i)$ of intersection $v_i$ for OVs is the sum of absolute pressures over all traffic movements, which can be defined as:
\begin{equation}
P_o(i) = \sum_{(l,l')\in i}{\big| x_o(l)-x_o(l') \big |}
\end{equation}
where $x_o(l)$ is the number of OVs on an entering lane $l$ and $x_o(l')$ is the number of OVs on an exiting lane $l'$.
What's more, considering the traffic priority of different types of vehicles, we utilize their proportions in the traffic flow to assign the weights in the reward function.
Then we define the reward $r_i$ as:
\begin{equation}
r_i = -\frac{L_e(i)}{\eta} - \frac{P_o(i)}{1-\eta}
\end{equation}
where ${L_e(i)}$ is the number of EVs on the entering lanes of intersection $v_i$ and $\eta$ is the proportion of EVs in all vehicles.

\subsubsection{\textbf{Dynamic Directed Graph }}
\label{subsubsec:DDG}
The dynamic directed graph helps capture the dynamic impacts of neighboring intersections due to real-time route planning.
We construct the road network as a graph in which the weight of each edge is calculated as the corresponding real-time road network distance $dis$ between two intersections.
Then, we get the top-$K$ relevant neighboring intersections $\mathcal{K}_i$ of intersection $v_i$ based on the dynamic relational distance.
Dynamic relational distance helps the current intersection pay more attention to the traffic flow at the upstream intersection when an EV will come from the upstream of the current intersection.
Specifically, according to the planned route, the relational distance between upstream intersections and current intersection is set smaller by assigning a relational factor $\delta$ to these intersections.
For an edge $e_{i,i+1}$ in the route of EV ${R}$, the relational distance $dis\_r(i,i+1)$ from intersection $v_i$ to $v_{i+1}$ is calculated as:
\begin{equation}
dis\_r(i,i+1) = dis(i,i+1) \cdot \delta
\end{equation}
where $dis(i,i+1)$ is the road network distance from intersection $v_i$ to $v_{i+1}$ and $\delta$ is the relational factor.


\begin{figure} 
\centering
\includegraphics[height=3.2cm,width=7.5cm]{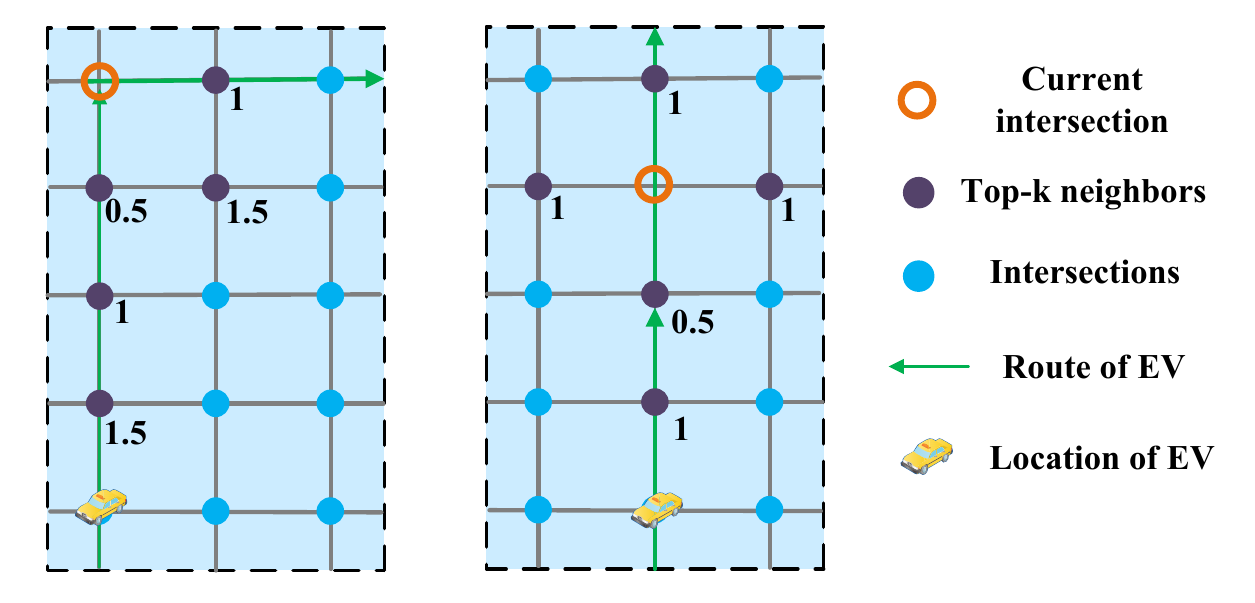}
\caption{Top-K relevant neighbor intersections based on dynamic relational distance. The black numbers denote the relational distances between current intersection and other intersections.}
\vspace{-10pt}
\label{top-k}
\end{figure}
Fig. \ref{top-k} illustrates the top-$K$ relevant neighboring intersections and the corresponding relational distance to the current intersection based on different routes of the EV. We set the road network distance between adjacent intersections as $1$, the discount factor $\delta$ = $0.5$ and $K$ = $6$. Please note that the current intersection itself is also included in the top-$K$ neighbors.

\subsubsection{\textbf{Multi-head attention relation kernel}}
\label{subsubsec:MHA}
The $m$-dimensional observation data $o_i$ of intersection $v_i$ are transformed into the $n$-dimensional hidden features $\mathcal{H}_i$ via a MLP:
\begin{equation}
\mathcal{H}_i = \sigma (o_iW_e + b_e)
\end{equation}
where $W_e$ and $b_e$ are the weight matrix and bias vector respectively.
Then we embed the representation of the current intersection $v_i$ and neighbor $v_j$ from the previous layer to get different types of importance score $e_{ji}$ of one neighbor.
Specifically, we utilize the multi-head attention mechanism where attention functions with different linear projections are performed in parallel to jointly attend to a neighbor from different representation subspaces with the following operations:
\begin{equation}
e_{ji}^h = (\mathcal{H}_iW_t^h) \cdot (\mathcal{H}_jW_s^h)^T
\end{equation}
where $h \in (1,2,\cdots,H)$ is the index of different representation subspaces and $e_{ji}^h$ is the importance score of neighbor $v_j$ to current intersection $v_i$ in the subspace $h$. Please note that $e_{ji}$ is usually different from $e_{ij}$ due to dynamically updated route planning.
We retrieve the general attention score between neighbors and the current intersection by normalizing the importance score of different neighbors in the same subspace: 
\begin{equation}
\alpha_{ji}^h = {\rm softmax} (e_{ji}^h)=
\frac{ {\rm exp} (e_{ji}^h/{\mu}) }
{ \sum_{j\in\mathcal{K}_i} { {\rm exp} (e_{ji}^h/{\mu})} }
\end{equation}
where $\mu$ is the temperature factor and $\mathcal{K}_i$ is the top-$K$ relevant neighboring intersections of intersection $v_i$.
Finally we model the overall influence of neighbors to the current intersection in different subspaces by combining the hidden feature representations $\mathcal{H}_j$ of all the top-$K$ relevant neighbors with their respective general attention scores $\alpha^h_{ji}$:
\begin{equation}
hm_i = \sigma\Big(W_q \cdot \big( \frac{1}{H} \sum_{h=1}^{H} \sum_{j \in \mathcal{K}_i} \alpha_{ji}^h (\mathcal{H}_j W_c^h) \big) + b_q \Big)
\end{equation}
The averaging operation of multi-head attention is one of the most feasible ways to conclude the neighborhood cooperation.

\subsubsection{\textbf{Centralized Critic Model}}
\label{subsubsec:phase_eval}
The key idea of RL is to utilize Bellman equation to estimate the long-term discounted cumulative reward of an action, which is significant for the transportation system with strong spatio-temporal correlations.
The long-term impact $\mathcal{R}$ of a signal control action is defined as follows:
\begin{equation}
\mathcal{R} = \sum_{t=0}^T \gamma^t r(\bm{o}_t^i,\bm{a}_t^i)
\end{equation}
where $r(\bm{o}_t^i,\bm{a}_t^i)$ is the immediate reward of action $\bm{a}_t^i$ based on the observation $\bm{o}_t^i$ at intersection $v_i$.
Based on the processed real-time observation information $hm_i$ , we leverage the deep RL to estimate the expected reward of the given state-action pair $(\bm{o}_t^i,\bm{a}_t^i)$ as $Q(\bm{o}_t^i,\bm{a}_t^i|\theta)$, which can be calculated as:
\begin{equation}
Q(\bm{o}_t^i) = {hm}_i W_p + b_p
\end{equation}
where $W_p \in \mathbb{R}^{c \times p}$ and $b_p$ are the training parameters, $p$ is the number of phases (action space) and $\theta$ represents all the trainable variables in our centralized critic model. The phase action with the maximum long-term reward will be chosen.
We optimize our control policy by minimizing the loss function as follows:
\begin{equation}
L(\theta) = \sum_{t=1}^{T} \sum_{i=1}^{I} \big( Q(\bm{o}_t^i,\bm{a}_t^i|\theta) - y_t  \big)^2
\end{equation}
where $T$ is the number of time steps, $I$ is the number of intersections and $y_t$ is the target $Q$ value defined as:
\begin{equation} 
y_t = r_t^i + \gamma \ \max_{\bm{a}_{t+1}}Q(\bm{o}_{t+1}^i,\bm{a}_{t+1}^i)
\end{equation}
\begin{figure*}[!t]
    \centering
    \vspace{-0.35cm}
    \subfigbottomskip=1.5pt
    \subfigure[Hefei dataset]{
      \label{fig_hefei_map}
      \includegraphics[height=1.5in]{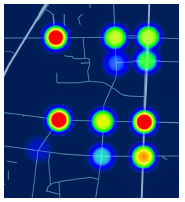}}
    \subfigure[Jinan dataset]{
      \label{fig_jinan_map}
      \includegraphics[height=1.5in]{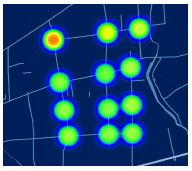}}
    \subfigure[Hangzhou dataset]{
      \label{fig_hangzhou_map}
      \includegraphics[height=1.5in]{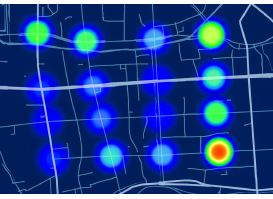}}
    \subfigure[NewYork dataset]{
      \label{fig_newyork_map}
      \includegraphics[height=1.5in]{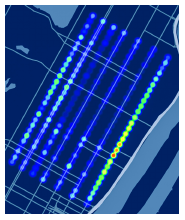}}
    \caption{The spatial distribution of traffic flows of different datasets during the experiment period. Each node on the road network represents the  traffic flow passing through an intersection. The red, yellow, green and blue colors indicate the decreasing traffic volume.}
    \label{space_distribution}\vspace{-10pt}
\end{figure*}
\subsection{Complexity analysis}
In this subsection, we analyze the scalability of \emph{LEVID\_Dy}, namely the RL part of \emph{LEVID}. 
Specifically, we suppose our model gets the $m$-dimensional input data and each layer has $n$ neurons; the scale of traffic signal phase space is $p$. 
The time and space complexities are analyzed based on the following assumptions: 
(a) all the distributed actors leverage the centralized critic model to predict the long-term discounted reward of a traffic signal action; 
(b) each target intersection gets the top-$K$ relevant neighbors based on a breadth first search with a total search number of 2$K$, as excessive search range may cause unnecessary computational consumption;
(c) the multi-head attentions are computed independently with the same time consumption as that of single-head attention, and the embedding process of either source or target intersection can be executed simultaneously;
(d) all the actors can execute the prediction process independently. 
Then the time complexity in each component is: (a) \emph{top-$K$ search}: $O(2K+KlogK)$; (b) \emph{MLP}: $O(mn)$; (c) \emph{Graph Attentional layers}: $O:(n^2+n^2)$; (d) \emph{Q-value Prediction layer}: $O(np)$. And the total time complexity is $O(n(m+2n+p)+K)$, which is approximately equal to $O(n^2)$.

As for space complexity, the size of weight matrix and bias vectors in each component are as follows:
(a) \emph{top-$K$ search}: $2K$
(b) \emph{Observation Embedding layer}: $mn+n$;
(c) \emph{Graph Attention layers}: $3n^2+(n^2+n) = n(4n+1)$;
(d) \emph{Q-value Prediction layer}: $np+p$.
Then the total number of parameters to store is $O(n(4n+m+p+2) + 2K + p)$.
Normally, the size of the hidden layer $n$ is far greater than that of the data dimension $m$ and state space $p$. Therefore, the space complexity of \emph{LEVID\_Dy} is approximately equal to $O(n^2)$. For a method with $N$ separate RL models (without parameter sharing) to control traffic signals in $N$ intersections, the space complexity is approximately equal to $O(n^2 \cdot N)$.

%% file: 6evaluation.tex
\section{Performance Evaluation}
\label{sec:6evaluation}
We conduct experiments with an open-source traffic simulator called \textit{CityFlow} \cite{zhang2019cityflow}. 
After the traffic trajectory data with specific route and start time are fed into the simulator, each OV moves towards its destination according to the environmental setting and the phase of traffic lights.
The simulator provides states to a traffic signal control strategy and performs traffic signal actions from the control strategy.
Meanwhile, we add a route planning module to the simulator, which controls an EV towards its destination along a dynamic route.


\subsection{Setting}
Both synthetic and real-world datasets are utilized to evaluate the effectiveness and efficiency of different approaches.
One synthetic dataset is used to generate uniform traffic flows to test the performance of various approaches in steady traffic conditions.
Four real-world traffic flow datasets are collected from four cities for evaluations on realistic and dynamic traffic conditions, and the road networks are imported to the simulator from OpenStreeMap\footnote{https://www.openstreetmap.org}.
We randomly select some vehicles in the traffic flows as EVs whose routes could be dynamically changed according to a route planning module, and the rest of vehicles still follow their original routes.
Indeed, EVs usually account for a very low proportion of the overall traffic flows in the real world.
Nevertheless, if the proportion of EVs is set too small, there will be only a small amount of transition experiences of EVs interacting with the environment, which will cause sparse rewards in the training of an RL model. 
Therefore, we set the proportion of EVs as $1\%$ for model training, which can not only generate necessary interaction experiences, but also simulate conflicted situations of multiple EVs at the same intersection.
While the proportion of EVs is set according to the real-world conditions for model testing (around $0.37\sim1$\textperthousand). Table \ref{table_data} lists the statistics of different datasets.
Fig. \ref{space_distribution} further shows the spatio-temporal distribution of traffic flows. The detailed descriptions on how we set or preprocess these datasets are as follows:

\begin{table}[!htbp]
\caption{Statistics of five datasets}
\label{table_data}
\begin{center}
\begin{tabular}{llllll} 
\hline
\multirow{2}{*}{Dataset} & \multirow{2}{*}{\# intersections} & \multicolumn{4}{c}{Arrival rate (vehicles/300s)}  \\
\multicolumn{2}{c}{} & Mean & Std & Max & Min \\ \hline  
$D_{Synthetic}$ & 36 & 97.5 & 0 & 97.5 & 97.5 \\ \hline
$D_{Hefei}$ & 11 & 437.92 & 51.11 & 514 & 341 \\ \hline
$D_{Jinan}$ & 12 & 457.83 & 46.22 & 544 & 363 \\ \hline
$D_{Hangzhou}$ & 16 & 513.75 & 242.34 & 875 & 203 \\ \hline
$D_{NewYork}$ & 196 & 879.34 & 315.14 & 1314 & 416 \\ \hline
\end{tabular}
\end{center}\vspace{-10pt}
\end{table}

\begin{itemize}[leftmargin=1em,itemindent=0pt,listparindent=0pt]
\item $D_{Synthetic}$: Following the setting of \cite{wei2019colight}, this dataset contains a $6 \times 6$ grid network where each intersection has 4 directions (West$\to$East, East$\to$West, South$\to$North, North$\to$South) and 3 lanes (300 meters long and 3 meters wide) for each direction. In the traffic flow, vehicles come uniformly with 300 vehicles/lane/hour in the East$\leftrightarrow$West direction and 90 vehicles/lane/hour in the South$\leftrightarrow$North direction.
\item $D_{Hefei}$: There are 11 intersections in one region of Baohe district, Hefei city, China. 
The traffic flow data are collected by roadside surveillance cameras during 9-11 a.m. on the working days of April 2021. 
The cameras record the time, location and vehicle ID. We set the traffic volume as the number of vehicles passing through these intersections for experiments.
\item $D_{Jinan}$\cite{wei2019colight}: There are 12 intersections in Dongfeng Sub-district, Jinan, China. 
The traffic flow data are collected by cameras in the similar way to $D_{Hefei}$.
\item $D_{Hangzhou}$\cite{wei2019colight}: There are 16 intersections in Gudang Sub-district, Hangzhou, China.  
The traffic flow data are collected by cameras in the similar way to $D_{Hefei}$.
\item $D_{NewYork}$\cite{wei2019colight}: There are 192 intersections in the Upper East Side of Manhattan. 
The traffic flow data are collected based on the taxi trip data containing the origin and destination geo-locations of each trip. 
The geo-locations are mapped to intersections and the corresponding shortest path between them is obtained. 
The trips falling within the selected areas are chosen for experiments.
\end{itemize}

\subsection{Compared methods} 
We compare our \emph{LEVID} approach with various baselines and variants of \emph{LEVID}. We summarize these approaches from two aspects in Table \ref{table_method}. On the one hand, considering the key technologies (e.g., vehicle-centric or road-centric, whether or not to use the RL method), they can be classified as: conventional traffic signal control approaches,  RL-based traffic signal control approaches, route planning approaches and cooperative vehicle-road scheduling approaches. On the other hand, they can also be classified according to whether an approach is specially designed for EVs or just for OVs. 
All RL-based approaches are learned without any pre-trained parameters for fair comparison. 
The evaluation metric is the \textbf{average travel time} of all the EVs or OVs between origin and destination (in seconds).
\newcommand{\tabincell}[2]{\begin{tabular}{@{}#1@{}}#2\end{tabular}} 

\begin{table}[!htbp]
\caption{Classification of various approaches}
\label{table_method}
\begin{center}
\setlength{\tabcolsep}{0.3mm}
\begin{tabular}{ccc} 
\hline
 & OVs & EVs  \\ \hline  
\tabincell{c}{Conventional Trafic \\ Signal Control} & \tabincell{c}{ \emph{FixedTime}, \emph{MaxPressure}} & \emph{GreenWave} \\ \hline
\tabincell{c}{RL-based Traffic \\ Signal Control} & \tabincell{c}{\emph{Individual RL}, \\ \emph{OneModel}, \emph{CoLight}} &  \tabincell{c}{\emph{LEVID\_UnDy}, \\ \emph{LEVID\_Dy}} \\  \hline 
\tabincell{c}{Route \\ Planning} & $/$ & \emph{Dijkstra} \\ \hline
\tabincell{c}{CVRS\\ (loosely coupled)} & $/$ & \emph{AAF} \\ \hline
\tabincell{c}{CVRS\\ (tightly coupled)} & $/$ & \tabincell{c}{\emph{LEVID\_APF}, \\ \emph{LEVID}} \\ \hline
\end{tabular}
\end{center} \vspace{-15pt}
\end{table}

\textbf{Baselines:}
\begin{itemize}[leftmargin=1em,itemindent=0pt,listparindent=0pt]
\item
\emph{FixedTime} \cite{koonce2008traffic}: It's the most commonly used traffic signal control method with preset offsets in the real world. It utilizes a pre-determined schedule plan considering the cycle length and phase time to handle the traffic flow.
\item
\emph{MaxPressure} \cite{lioris2016adaptive}: It's the most popular network-level traffic signal control approach in the transportation field, which greedily selects the phase with the maximum pressure.
\item
\emph{GreenWave} \cite{kang2014traffic}: It allows all the traffic lights in the route to turn green so that EVs can pass intersections continuously along the emergency corridor.
All the intersections share the same green phase length for each movement.
\item
\emph{Dijkstra} \cite{chen2014path}: It's a vehicle-centric scheduling approach, which builds a dynamic road network model for vehicles evacuation based on the Dijkstra algorithm.
\item
\emph{AAF} \cite{djahel2015reducing}: 
It's an advanced adaptive and fuzzy approach to reduce emergency services response time. It selects the fastest path for an EV in advance and gives priority to the EV as soon as it approaches the traffic lights on the preset route. Note that, it is just a \textit{loosely coupled} cooperation, as the route planning and traffic signal control modules are sequentially conducted, while our \textit{LEVID} has a \textit{tightly coupled} cooperation as the two modules are simultaneously conducted.
\item
\emph{Individual RL} \cite{wei2018intellilight}: It's the individual deep RL approach without considering the information of neighbors.
Each intersection is controlled by one heterogeneous agent which updates its own network independently. 
\item
\emph{OneModel} \cite{chu2019multi}: It designs the state and reward of the agent in the same way with \emph{Individual RL}. Each agent only considers the state of the roads connecting the controlled intersections and all agents share the same centralized critic model.
\item
\emph{CoLight} \cite{wei2019colight}: It's an RL-based traffic signal control approach utilizing graph attention networks to automatically extract traffic features of adjacent intersections for facilitating communication.
\end{itemize}

\textbf{Variants of \emph{LEVID}:}
\begin{itemize}[leftmargin=1em,itemindent=0pt,listparindent=0pt]
\item
\emph{LEVID\_UnDy}: It removes the real-time route planning module from the \emph{LEVID}. Meanwhile, its traffic signal control module removes the design of dynamic directed graph, and chooses the top-$K$ relevant neighbors based on the fixed geographic distance. This variant can show the improvements brought by the design of our state and reward function.
\item
\emph{LEVID\_Dy}: It removes the real-time route planning module from the \emph{LEVID} and retains the traffic signal control module, which selects the top-$K$ relevant neighbors based on a dynamic directed graph.
\item
\emph{LEVID\_APF}: It utilizes the artificial potential field method which only considers the gravity and immediate repulsion, instead of the real-time path planning module of \emph{LEVID}.
\end{itemize}

\begin{table*} 
\caption{Comparisons of average travel time of both OVs and EVs on the five datasets.}
\label{table_result}
\begin{center}
\setlength{\tabcolsep}{3.3mm}{
\begin{tabular}{c|cc|cc|cc|cc|cc}
\hline
\multirow{2}{*}{Model} & \multicolumn{2}{c|}{$D_{Synthetic}$}  & \multicolumn{2}{c|}{$D_{Hefei}$} & \multicolumn{2}{c|}{$D_{Jinan}$} & \multicolumn{2}{c|}{$D_{Hangzhou}$} & \multicolumn{2}{c}{$D_{NewYork}$}\\

{} & OVs & EVs  & OVs & EVs & OVs & EVs & OVs & EVs & OVs & EVs \\ \hline
FixedTime & 209.7 & 209.1 & 1175.9 & 1097.6 & 867.7 & 869.3 & 654.6 & 645.5 &  2239.1 & 1399.4\\
MaxPressure & 194.5 & 192.4 & 660.2 & 642.45  & 387.4 & 394.7 & 514.2 & 523.4 & 1666.1 & 1113.5\\
GreenWave & 203.8 & 169.6 & 1384.9 & 529.2  & 832.2 & 245.9 & 796.8 & 336.4 & 2497.1 & 611.6 \\
Dijkstra & 210.4 & 209.8 & 1247.9 & 1114.7 & 893.5 & 841.1 & 690.9 & 572.4 & 2020.5 & 1152.7 \\
AAF & 205.2 & 167.2 & 1417.8 & 544.3 & 829.3 & 241.7 & 676.7 & 343.1 & 2107.5 & 581.3 \\
Individual RL & \textbf{189.36} & 188.5  & \textbf{569.3} & 570.6 & 343.1 & 347.3 & \textbf{404.3} & 403.7 & $*$ & $*$\\
OneModel & 211.8 & 211.2 & 1411.3 & 1459.2 & 724.9 & 725.8 & 570.8 & 520.5 & 1979.1 & 1219.8\\
CoLight & 192.5 & 188.4 & 625.7 & 603.3 & \textbf{293.3} & 291.6 & 534.4 & 504.82 & 1459.5 & 906.6\\
LEVID\_UnDy  & 197.4 & 160.6 & 667.2 & 549.6 & 354.6 & 254.1  & 567.9 & 339.9 & 1596.3 & 732.8\\
LEVID\_Dy  & 193.1 & 157.8 & 674.3 & 490.5 & 352.5 & 235.6  & 586.7 & 307.7 & 1574.5 & 624.2\\
LEVID\_APF & 198.6 & 158.5 & 664.0 & 507.5 & 347.4 & 248.4 & 556.6 & 330.5 & 1434.7 & 667.9 \\
LEVID & 195.4 & \textbf{155.4} & 654.0 & \textbf{443.9} & 341.8 & \textbf{220.2} & 571.7 & \textbf{291.1} & \textbf{1431.8} & \textbf{546.5} \\ \hline
\end{tabular}}
\end{center}
\end{table*}

\subsection{Overall Performance Comparison} 
Table \ref{table_result} compares the average travel time of both OVs and EVs achieved by \emph{LEVID} and various baselines/variants on the five datasets.

\subsubsection{Advantages of \emph{LEVID} over Conventional Traffic Signal Control Approaches} From Table \ref{table_result}, we observe that \emph{FixedTime} has similar performance to \emph{MaxPressure} on $D_{Synthetic}$ with a uniformly simulated traffic flow, while \emph{MaxPressure} performs much better than \emph{FixedTime} on the four real-world datasets, indicating that \emph{MaxPressure} has a stronger ability of handling dynamic traffic flows.
However, both \emph{FixedTime} and \emph{MaxPressure} do not consider the priorities of EVs, resulting in that OVs and EVs have similar performance.
By contrast, \emph{GreenWave} is specially designed for EVs and greatly reduce the average travel time of EVs on all the datasets (at most 71.71\% and 45.07\% reductions compared with \emph{FixedTime} and \emph{MaxPressure}, respectively).
Nevertheless, \emph{GreenWave} increases the average travel time of OVs (at most 258.0s longer than \emph{FixedTime}), and there is a large performance gap between OVs and EVs (at most 1885.5s difference).
By contrast, our \textit{LEVID} and also its three variants not only greatly reduce the average travel time of EVs on all the datasets (at most 74.71\% and 50.94\% reductions compared with \emph{FixedTime} and \emph{MaxPressure}, respectively, by \textit{LEVID}), but also shorten the average travel time of OVs in most cases (at most 807.3s difference by \textit{LEVID}).

not deteriorate OVs too much (at most 125.8s difference by \textit{LEVID}).
The results demonstrate the obvious advantages of our \textit{LEVID} from two aspects: 1) utilize a learning-based traffic signal control strategy instead of a rule-based strategy for handling dynamic traffic flows, and 2) integrate it with a route planning strategy to further reduce the waiting time at intersections with a heavy traffic.


\subsubsection{Advantages of \emph{LEVID} over RL-based Traffic Signal Control Approaches}
From Table \ref{table_result}, we observe that \emph{Individual RL} performs better than other two RL-based baselines, \textit{OneModel} and \textit{CoLight}, and even performs best for OVs on four small-scale datasets, $D_{Synthetic}$, $D_{Hefei}$, $D_{Jinan}$ and $D_{Hangzhou}$.
It is because \emph{Individual RL} trains an exclusive agent for each intersection, which can evaluate the intersection state more accurately and reduce the performance loss.
However, \emph{Individual RL} cannot be applied to a large-scale dataset (e.g., $D_{NewYork}$) due to the low computational efficiency.
On the contrary, \emph{OneModel} utilizes a shared centralized critic network, which may ignore the differences between individuals, resulting in an inevitable performance loss.
Compared with \emph{OneModel}, \emph{CoLight} reduces the average travel time of EVs by $9.1\%$, $55.7\%$, $59.5\%$,  $6.4\%$ and $26.3\%$ on $D_{Synthetic}$, $D_{Hefei}$, $D_{Jinan}$, $D_{Hangzhou}$ and $D_{NewYork}$, respectively, because it considers the state of neighboring intersections and leverages the GAT to model the interactions between neighboring intersections.
Compared with \emph{CoLight}, our \emph{LEVID} and also its three variants achieve consistent and obvious performance improvements for EVs, and achieve similar performance for OVs.
More specifically, \emph{LEVID\_UnDy} reduces the average travel time of EVs by 14.76\%, 8.91\%, 12.94\%, 32.72\% and 19.24\% than \emph{CoLight} on the five datasets, respectively, which demonstrates the importance of the reward design considering both EVs and OVs simultaneously.
\emph{LEVID\_Dy} further reduces the average travel time of EVs by 1.70\%, 10.76\%, 7.36\%, 9.51\% and 14.82\% than \emph{LEVID\_UnDy} on the five datasets, respectively, which demonstrates the importance of utilizing a dynamic directed graph.
Moreover, our proposed approaches have a larger advantage with the increase of the road network scale. 

 \begin{figure*}[!t]
     \centering
     \vspace{-0.35cm}
     \subfigbottomskip=1.5pt
     \subfigure[Convergence speed on $D_{Synthetic}$]{
       \label{fig_synthetic}
       \includegraphics[width=3.1 in]{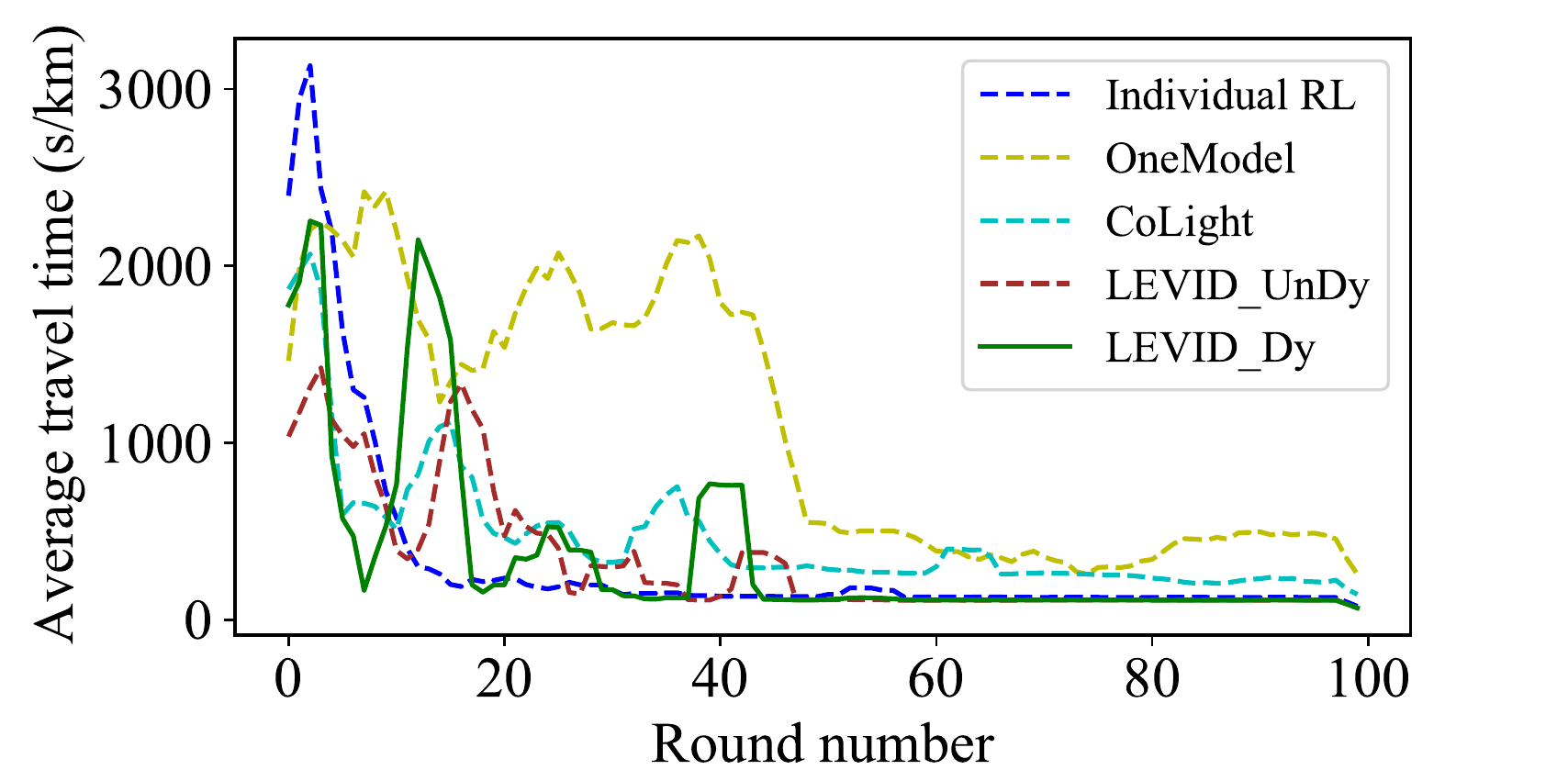}}
     \quad
     \subfigure[Convergence speed on $D_{Hefei}$]{
       \label{fig_hefei}
       \includegraphics[width=3.1 in]{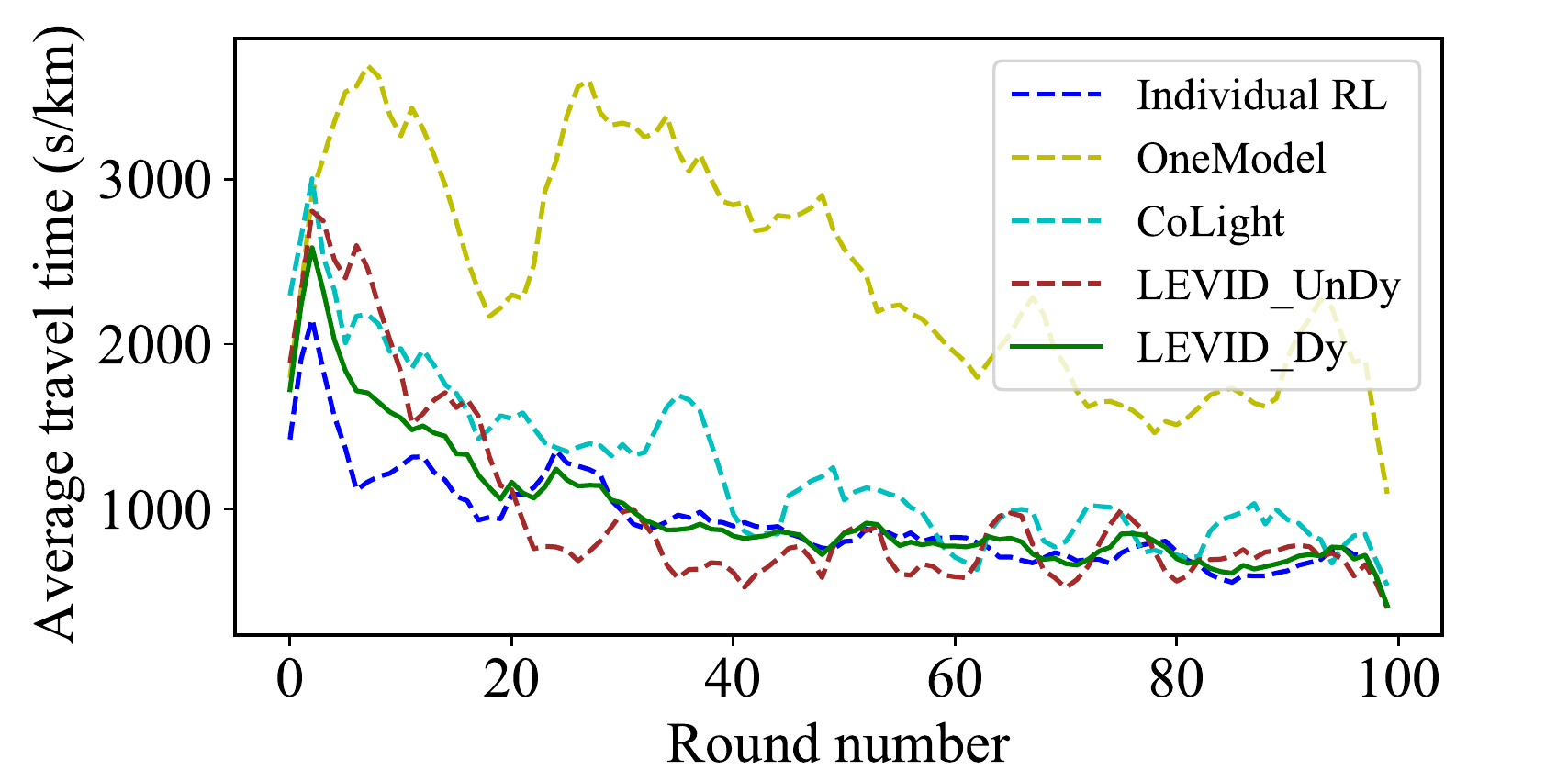}}
    
     \subfigure[Convergence speed on $D_{Jinan}$]{
       \label{fig_jinan}
       \includegraphics[width=3.1 in]{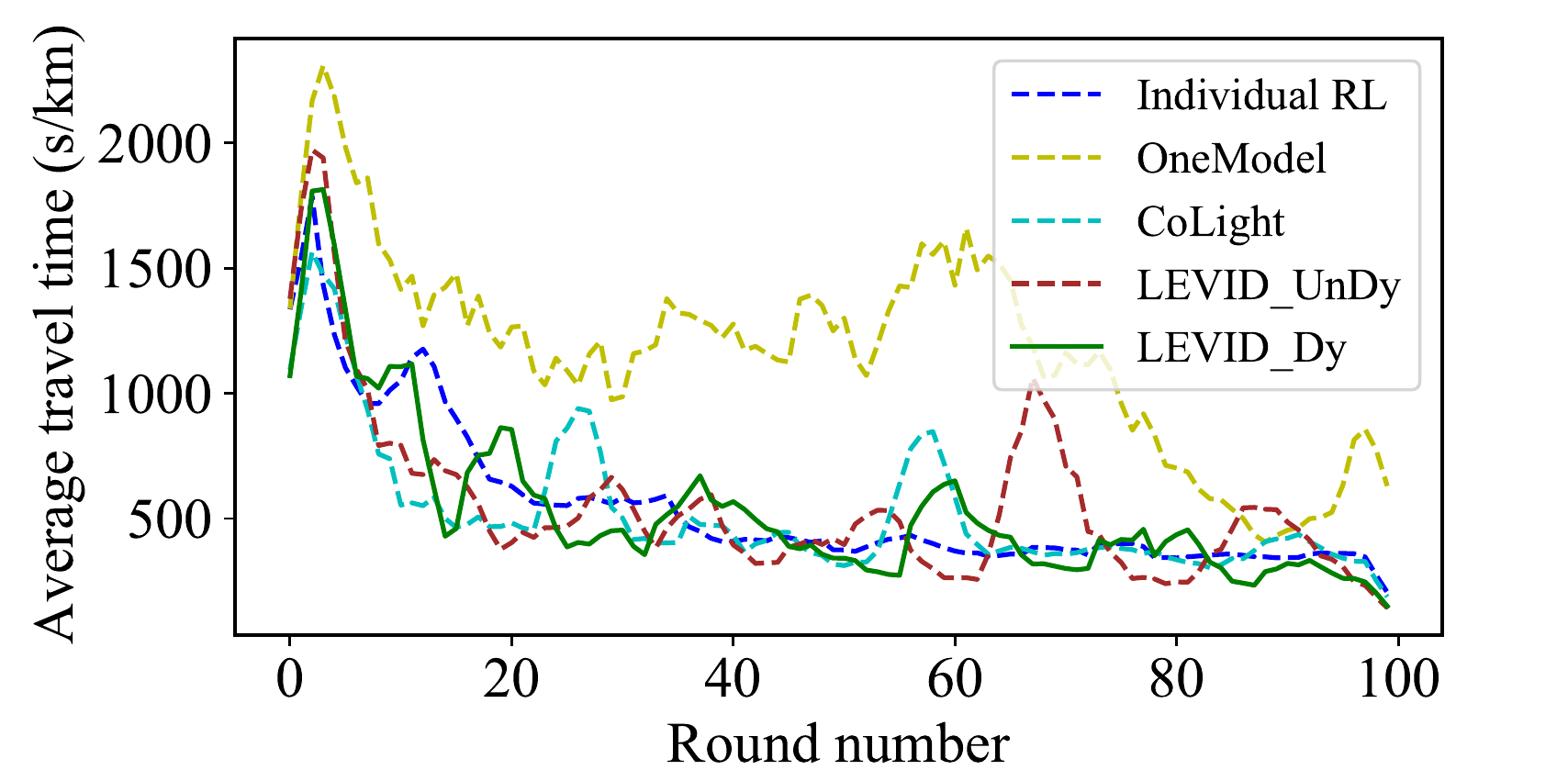}}
     \quad
     \subfigure[Convergence speed on $D_{Hangzhou}$]{
       \label{fig_hangzhou}
       \includegraphics[width=3.1 in]{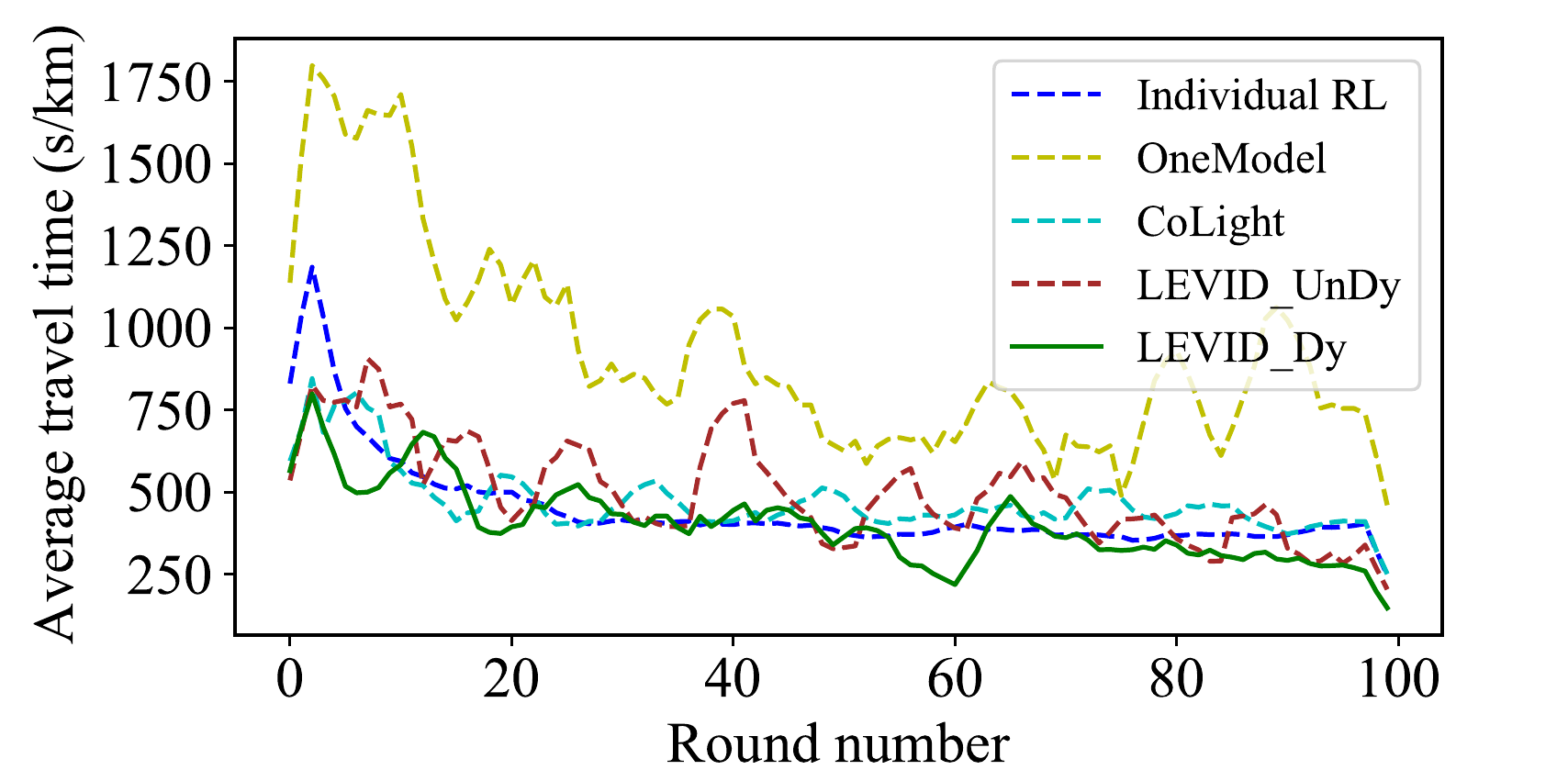}}
      
     \subfigure[Convergence speed on $D_{NewYork}$]{
       \label{fig_newyork}
       \includegraphics[width=3.1 in]{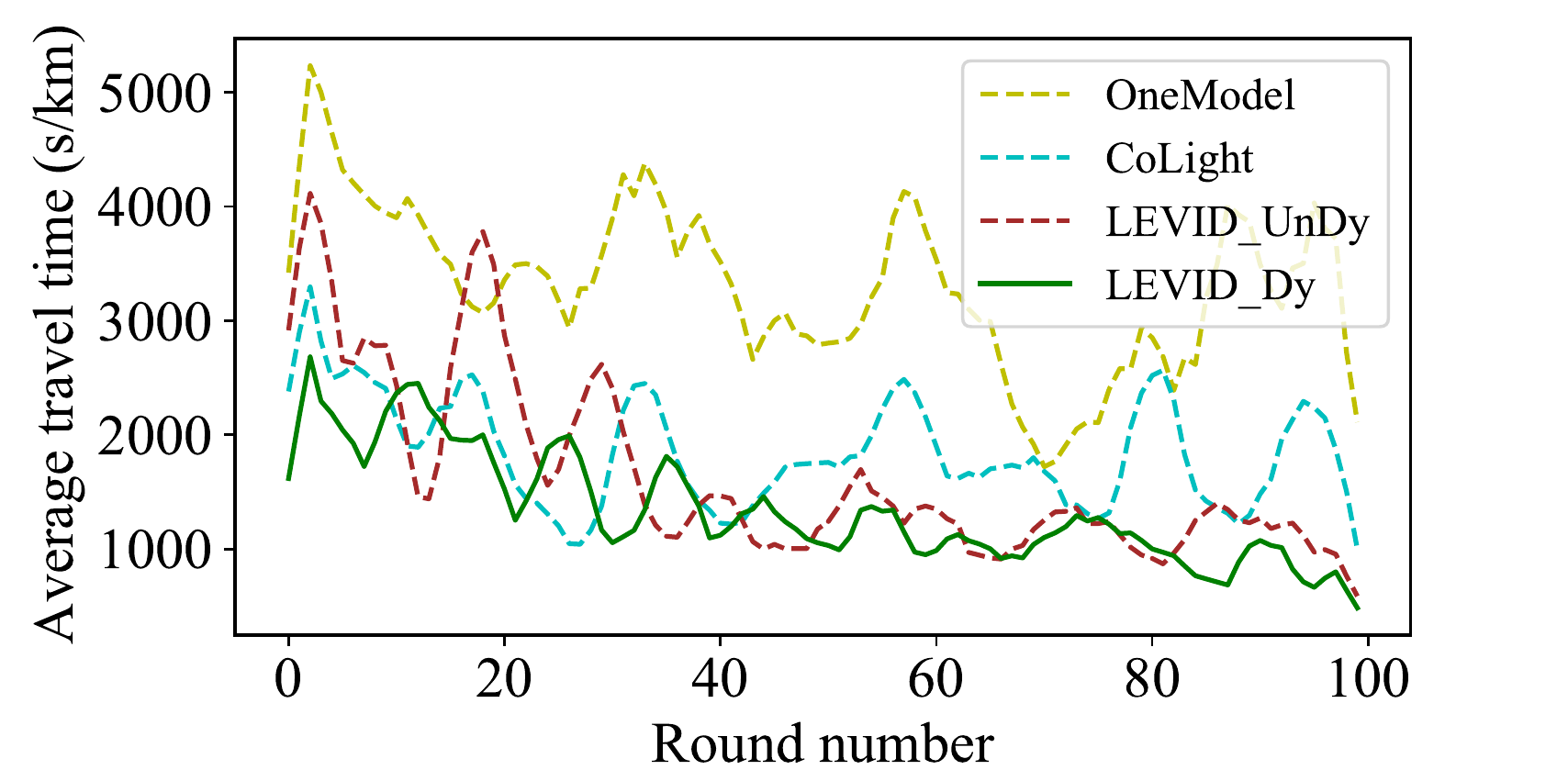}} 
     \quad
     \subfigure[Running time]{
       \label{runtime}
       \includegraphics[width=3.1 in]{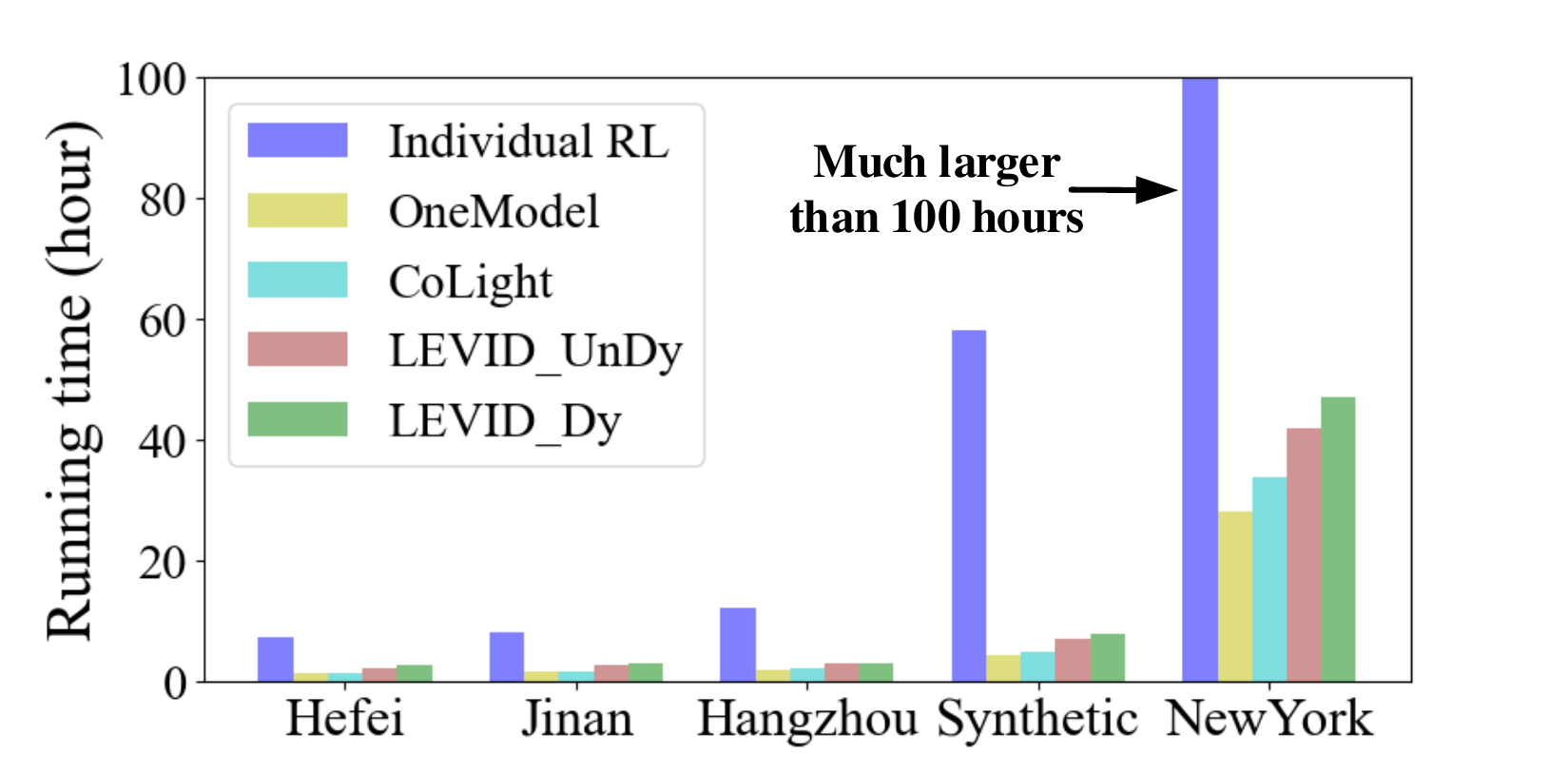}}  
     \caption{Convergence speed and running time of \emph{LEVID\_Dy} (green continuous curves) and other 4 RL-based traffic signal control approaches (dashed curves) during training.
     In most cases, \emph{LEVID\_Dy} starts with the best performance (Jumpstart), reaches to the pre-defined performance the fastest (Time to Threshold), and ends with the optimal policy (Aysmptotic). Curves are smoothed with a moving average of 5 points. 
     Note that, the convergence curve of \emph{Individual RL} is not provided on $D_{NewYork}$, as it cannot be applied to a large-scale dataset due to the low computational efficiency.
     }\vspace{-10pt}
     \label{fig_experiment}
 \end{figure*}

\subsubsection{Advantages of integrating a real-time route planning module}
From Table \ref{table_result}, we observe that \emph{Dijkstra} reduces the average travel time of EVs by 0.29\%, $10.72\%$, $5.86\%$, $17.25\%$ and $42.92\%$ than OVs in the same environment on the five datasets, respectively, which demonstrates the importance of route planning.
Nevertheless, this performance improvement is far less than that by several CVRS, because a vehicle-centric approach just avoids congested roads in a passive way while failing to proactively improve traffic conditions.
By contrast, \emph{AAF} reduces the average travel time of EVs by 18.53\%, 61.61\%, 70.97\%, 49.37\% and 72.36\% than OVs in the same environment on the five datasets, respectively.
Compared with \emph{GreenWave}, \emph{AAF} greatly reduces the average travel time of OVs due to the merit of integrating a route planning module.
However, \emph{AAF} cannot achieve consistent advantages for EVs especially on small-scale datasets, demonstrating its limited ability of handling frequently changing traffic flow in the way of planning routes in advance.
Compared with \emph{AAF}, \emph{LEVID\_APF} reduces the average travel time of OVs and EVs by at most 58.19\% and 6.81\%, respectively, on the five datasets, which demonstrates the importance of the tightly coupled cooperation between route planning and traffic signal control.
In spite of this, we observe that \emph{LEVID\_APF} performs worse than \emph{AFF} for EVs on $D_{NewYork}$, because \emph{LEVID\_APF} does not consider the limitation of the immediate repulsion in a large-scale road network.
By contrast, \emph{LEVID} reduces the average travel time of EVs by 1.96\%, 12.54\%, 11.48\%, 11.94\% and 18.26\% than \emph{LEVID\_APF} on the five datasets, respectively, which demonstrates the importance of considering the long-term repulsion.

\subsection{Convergence comparison}
In Fig. \ref{fig_experiment}, we present the average travel time of EVs evaluated at each episode to the corresponding learning curves for the five RL-based traffic signal control approaches. The results show that our method has better performance in both \emph{time to threshold} (learning time to achieve a pre-specified performance level) and \emph{asymptotic performance} (final learned performance). The convergence curve also presents the influence of dynamic traffic flows on the convergence. The convergence curves on the synthetic dataset are smoother while the dynamic real-world datasets bring some fluctuations to the convergence curve of most RL-based approaches.
The training time (total time for 100 episode training) of all RL-based traffic signal control approaches are also presented. 
For fair comparison, each model is trained individually. 
As shown in Fig. \ref{runtime}, the time consumption of \emph{LEVID\_Dy} is much less than that of \emph{Individual RL} and all the approaches with centralized model are efficient. 
This is consistent with the complexity analysis of the centralized model.
In addition, the actual time consumption of \emph{Individual RL} on $D_{NewYork}$ is not provided, as each episode takes more than 100 hours when all models are trained centrally on one server.

\begin{figure} 
\centering
\includegraphics[width=3.1 in]{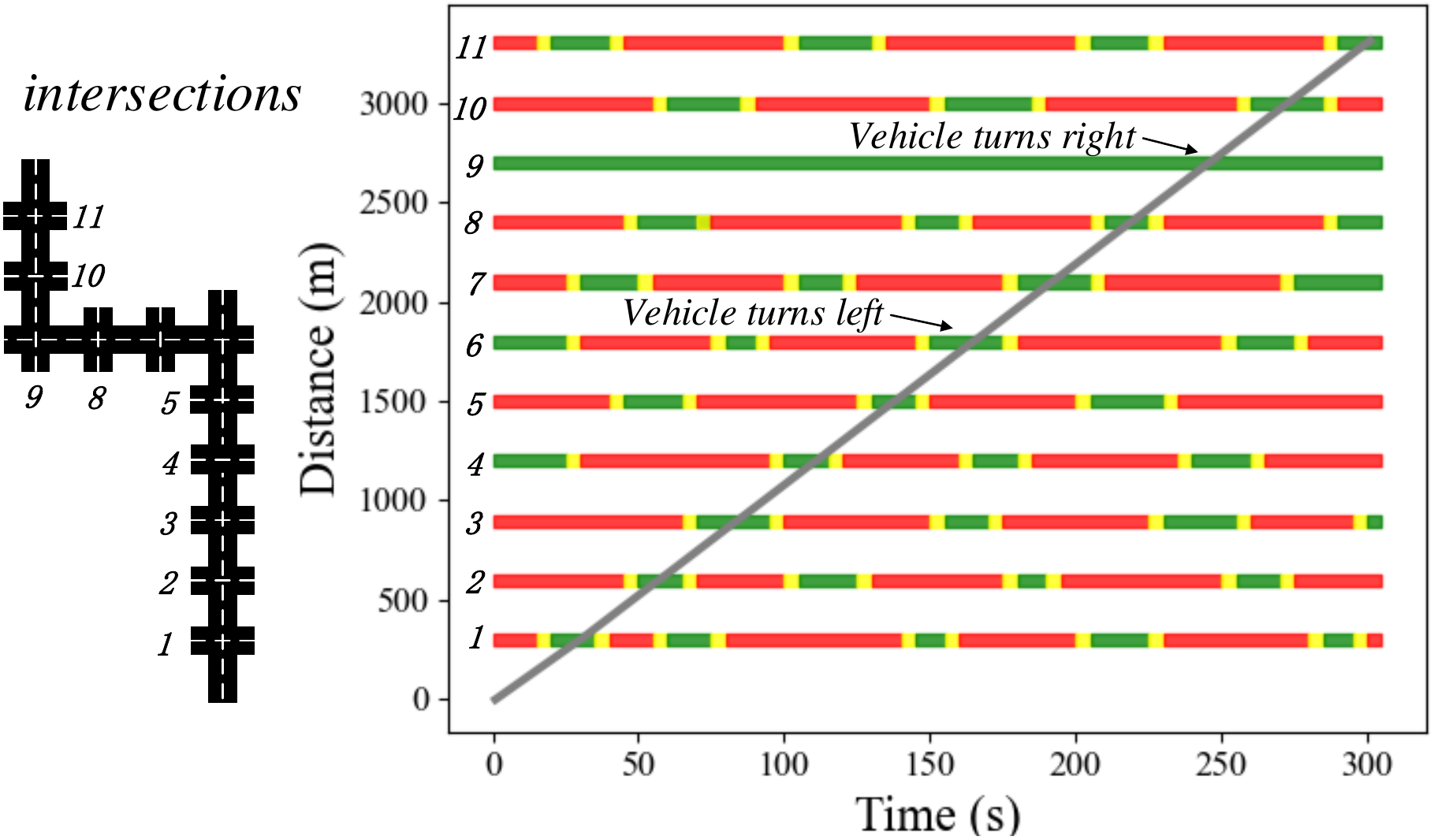}  
\vspace{-8pt}
\caption{Space-time diagram with signal timing plan to illustrate the effect of learned coordination strategy on dataset $D_{NewYork}$.
}
\vspace{-15pt}
\label{casestudy}
\end{figure}

\subsection{Case study}
The time-space diagrams are utilized to show the trajectory of one EV and the corresponding traffic signal control plan on dataset $D_{NewYork}$.
In Fig. \ref{casestudy}, the left part shows the real-world network structure and the right part denotes the specific driving process of the EV.
The x-axis is the time and y-axis denotes the distance.
There are one gray line denoting the trajectory of the EV and $11$ bands with green-yellow-red colors denoting the changing phases of $11$ intersections in this trajectory.
This EV turns right at the 9th intersection, where the right turn signal is always green.
Fig. \ref{casestudy} illustrates that the EV takes $300s$ to pass through $11$ intersections. We can observe that \emph{LEVID} can automatically form a green wave for the EV to help it pass quickly.

%% file: 8relatedwork.tex
\vspace{-10pt}
\section{Related Work}
\label{sec:2relatedwork}

\subsection{Vehicle-centric Scheduling}
The \textit{vehicle-centric scheduling} methods aim at scheduling vehicles with the best routes that can minimize the travel cost or satisfy personalized preferences.
Most of studies focus on route planning for OVs, which can be broadly divided into two categories, \textit{cost-centric} routing \cite{yang2014stochastic,hu2018risk,yuan2019weight,li2019time, pedersen2020fast,yang2018pace, pedersen2020hybrid} and \textit{trajectory-based} routing \cite{yuan2010t, dai2015personalized, guo2018learning, yang2020learning}.
Some \textit{cost-centric} studies mainly focus on route planning on a dynamic stochastic graph with time-dependent, uncertain edge weights \cite{yang2014stochastic,hu2018risk,yuan2019weight,li2019time, pedersen2020fast}. 
Other \textit{cost-centric} studies \cite{yang2018pace, pedersen2020hybrid} take the dependencies among time distributions of different roads into account to improve the accuracy of travel time estimation.
The \textit{trajectory-based} studies focus on leveraging historical trajectories for path recommendation \cite{yuan2010t, dai2015personalized, guo2018learning, yang2020learning}. 
However, these methods may not apply to EVs as the main concerns of EVs should be the time sensitivity rather than the personalized preferences.

Only a few early studies focus on route planning for EVs \cite{nordin2012finding,chen2014path,barrachina2014reducing}.
Nordin et al. \cite{nordin2012finding} utilize A* algorithm to determine the shortest path for dispatching an ambulance to a specific ambulance station or emergency site.
Chen et al. \cite{chen2014path} analyze three different emergency evacuation cases and build a dynamic road network model for vehicles evacuation based on the Dijkstra algorithm.
Barrachina et al. \cite{barrachina2014reducing} utilize vehicular communications to accurately estimate the traffic density in a certain area and help reduce the emergency services arrival time with evolution strategies.
However, these methods just avoid congested roads in a passive way, while failing to proactively improve the traffic condition to shorten the travel time of EVs.
\subsection{Road-centric Scheduling}
The road-centric scheduling methods aim at actively improving traffic conditions by traffic signal control technologies \cite{xu2019exploring, sommer2010bidirectionally}.
Extensive studies focus on traffic signal control for OVs, and the mainstream technologies have undergone an development from rule-based methods to learning-based methods \cite{wei2018intellilight, van2016coordinated, wei2019presslight, chen2020toward, wang2020stmarl}.
The conventional traffic signal control method Maxpressure \cite{lioris2016adaptive, varaiya2013max} measures the traffic flow in real time and changes the current phase according to the rule-based preset scheme.
Reinforcement learning based methods attempt to address traffic signal control problem by interacting with the environment and learning from real-time data. Some studies leverage tabular Q-learning \cite{abdulhai2003reinforcement, el2010agent} and deep reinforcement learning \cite{wei2018intellilight} for the traffic signal control of single intersection.
For the multi-intersection traffic signal control, the centralized RL method \cite{van2016coordinated, kuyer2008multiagent} models the actions of all agents jointly and negotiates the traffic signal control with centralized optimization, which is computationally expensive. While the decentralized RL method makes its decision based on observation of each independent agent.
Some methods \cite{nowe2012game, chen2020toward, arel2010reinforcement, el2013multiagent, zang2020metalight} 
handle the non-stationary impacts of other agents in complicated environment with exquisite reward design, which requires more human expert experience. In contrast, other methods \cite{nishi2018traffic, zheng2019diagnosing, wang2020stmarl, wei2019colight} add neighbors' traffic condition into observation and enable agents to behave as a group and form coordination. However, these methods do not consider the priority of EVs. 

By contrast, the existing traffic signal control studies for EVs are still limited to the rule-based methods \cite{kang2014traffic,  younes2018efficient, rosayyan2020decentralized}.
Kang et al. \cite{kang2014traffic} propose an EV signal coordination approach to provide ``green wave'' for EVs. 
Younes et al. \cite{younes2018efficient} design a real-time dynamic traffic signal control method which can handle the presence of one or more EVs over the road networks.  
Rosayyan et al. \cite{rosayyan2020decentralized} leverage a global navigation satellite system based on geo-fencing techniques to identify the entry of EVs and provide green signal automatically.
However, these rule-based methods rarely consider the impact of scheduling strategy on OVs and cannot interact with the environment in real-time.


\subsection{Cooperative Vehicle-Road Scheduling}
The cooperative vehicle-road scheduling provides route planning and traffic light control for EVs simultaneously.
Djahel et al. \cite{djahel2015reducing} design a traffic signal controller, which finds the quickest path for an EV in advance, and utilize RFID to give priority to an EV as soon as it approaches the traffic lights on this route. Karmakar et al. \cite{karmakar2020smart} determine the signal lights to be green based on the current traffic condition and calculate the priority levels of different EVs based on the type and the severity of an incident in case of the conflict between EVs. This work also considers the impact on the traffic in the neighboring roads surrounding the EV’s travel route.
However, these methods are mainly based on pre-set fixed rules and simplified assumptions. They cannot be updated synchronously as the real-time dynamic traffic flow changes \cite{humagain2020systematic}.

%% file: 9conclusion.tex
\section{Conclusion}
\label{sec:9conclusion}
In this paper, we consider a cooperative vehicle-infrastructure system to help EVs arrive faster.
Based on the key insight that real-time vehicle-road information interaction and strategy coordination can bring more benefits, we propose \emph{LEVID}, a learning-based cooperative vehicle-road scheduling approach. \emph{LEVID} contains a real-time route planning module and a collaborative traffic signal control module, which influences each other and makes decisions iteratively. The first module adapts the artificial potential field method to handle the real-time changes of traffic signals and jump out of the local optimum. The second module utilizes the multi-agent reinforcement learning framework to handle the traffic features that are hard to be combined linearly based on human experience and predefined rules. It further leverages graph attention networks based on a dynamic directed graph to model the interactions between intersections. Extensive experiments based on multiple real-world datasets demonstrate that our approach outperforms the state-of-the-art baselines.